\documentclass[jair,twoside,11pt,theapa]{article}
\usepackage{jair,theapa,epsfig,latexsym, verbatim}
\usepackage{epstopdf}
 \newcommand{\postscript}[2]
    {\setlength{\epsfxsize}{#2\hsize}
    \centerline{\epsfbox{#1}}}

\jairheading{32}{2008}{825--877}{06/07}{08/08}
\ShortHeadings{Qualitative System Identification}
{Coghill, Srinivasan, \& King}
\firstpageno{825}

\begin{document}
	\title{Qualitative System Identification from Imperfect Data}

	\author{\name George\,M. Coghill \email g.coghill@abdn.ac.uk \\
		\addr School of Natural and Computing Sciences \\
			University of Aberdeen, Aberdeen, AB24 3UE. UK.
		\AND
	        \name Ashwin Srinivasan \email ashwin.srinivasan@in.ibm.com \\
		\addr IBM India Research Laboratory \\
			4, Block C, Institutional Area \\
			Vasant Kunj Phase II, New Delhi 110070, India. \\
			and \\
			Department of CSE and Centre for Health Informatics \\
			University of New South Wales, Kensington \\
			Sydney, Australia.
		\AND
	        \name Ross\,D. King \email rdk@aber.ac.uk \\
		\addr Deptartment of Computer Science \\  University of Wales, Aberystwyth, SY23 3DB.  UK.}

	\maketitle

\renewcommand{\floatpagefraction}{0.9}
\renewcommand{\dblfloatpagefraction}{0.9}

\begin{abstract}
Experience in the physical sciences suggests that the only realistic means of understanding complex systems is through the use of mathematical models. Typically, this has come to mean the identification
of quantitative models expressed as differential equations. Quantitative modelling works best when the structure of the model (i.e., the form of the equations) is known; and the primary concern is one of estimating the values of the parameters in the model. For complex biological systems, the model-structure is rarely known and the modeler has to deal with both model-identification
and parameter-estimation. In this paper we are concerned with providing automated assistance to the first of these problems. Specifically, we examine the identification by machine of the structural relationships between experimentally observed variables. These relationship will be expressed in the
form of qualitative abstractions of a quantitative model. Such qualitative models may not only provide clues to the precise quantitative model, but also assist in understanding the essence of  that model. Our position in this paper is that background knowledge incorporating system modelling principles can
be used to constrain effectively the set of good qualitative models. Utilising the model-identification framework provided by Inductive Logic Programming (ILP) we present empirical support for this position using a series of increasingly complex artificial datasets. The results are obtained with qualitative and
quantitative data subject to varying amounts of noise and different degrees of sparsity.  The results also point to the presence of a set of qualitative states, which we term {\it kernel subsets\/}, that may be necessary for a qualitative model-learner to learn correct models. We demonstrate scalability of the method to biological system modelling by identification of the glycolysis metabolic pathway from data.
\end{abstract}

%-----------------------------------------------------------------
\section{Introduction}

There is a growing recognition that research in the life sciences will increasingly be concerned with ways of relating large amounts of biological and physical data to the structure and function of
higher-level biological systems. Experience in the physical sciences suggests that the only realistic
means of understanding such complex systems is through the use of mathematical models.  A topical example is provided by the Physiome Project which seeks to utilise data obtained from sequencing
the human genome to understand and describe the human organism using models that:
``\ldots include everything from diagrammatic schema, suggesting relationships among elements composing a system, to fully quantitative, computational models describing the behaviour of the physiological systems and an organism's response to environmental change'' (see {\tt http://www.physiome.org/}).   This paper is concerned with a computational tool that aims to assist in the identification of mathematical models for such complex systems.

Broadly speaking, system identification can be viewed as ``the field of modelling dynamic systems from experimental data'' \cite{soderstrom:sysid}. We can distinguish here between: (a) ``classical'' system identification techniques, developed by control engineers and econometricians; and (b) machine learning techniques, developed by computer scientists. There are two main aspects to this activity. First, an appropriate structure has to be determined (the model-identification problem). Second, acceptably accurate values for parameters in the model are to be obtained (the parameter-estimation problem).  Classical system identification is usually (but not always) used when the possible model structure is known {\em a priori\/}.  Machine learning methods, on the other hand, are of interest when little
or nothing is known about the model structure. The tool described here is a machine learning technique that identifies \textit{qualitative} models from observational data. Qualitative models are non-parametric; therefore  all the computational effort is focussed on model-identification (there are no parameters to be estimated). The task is therefore somewhat easier than more ambitious machine learning programs that attempt to identify parametric quantitative models \cite{Bradley00,Dzeroski92,Dzeroski95,Todorovski00}. Qualitative model-learning has a number of other advantages: the models are quite comprehensible; system-dynamics can be obtained relatively easily; the space of possible models is finite; and noise-resistance is fairly high. On the down-side, qualtitative model-learners have often produced models that are under- or over-constrained; the models can only provide clues to the precise mathematical structure; and the models are largely restricted to being abstractions of ordinary differential equations (ODEs).
We attempt to mitigate the first of these shortcomings by adopting the framework of Inductive Logic Programming (ILP) \cite<see>{BergadanoF96,mugg:der}. Properly constrained models are identified
using a library of syntactic and semantic constraints---part of the {\it background knowledge} in the ILP system---on physically meaningful models. Like all ILP systems, this library is relatively easily extendable.  Our position in this paper is that: 

\begin{quote}
{\it Background knowledge incorporating physical (and later, biological) system modelling principles
can be used to constrain the set of good qualitative models.\/}
\end{quote}

\noindent
Using some some classical physical systems as test-beds we demonstrate empirically that: 

\begin{itemize}
\item[--] A small set of constraints, in conjunction 	with a Bayesian scoring
	function, is sufficient to identify correct models.
\item[--] Correct models can be identified from qualitative or quantitative data which need not contain measurements 	for all variables in the model; and they can be learned with sparse 	data with large amounts of noise. 	That is, the correct models can be identified when the input data are incomplete, incorrect, or both.
\end{itemize}

\noindent
A closer examination  of the performance on these test systems has led to the discovery of what we term {\it kernel subsets\/}: minimal qualitative states that when present guarantee our implementation will
identify a model correctly. This concept may be of value to other model identification systems.

Our primary interests, as made clear at the outset, lie in biological system identification. The completion of the sequencing of the key model genomes and the rise of new technologies have opened up the prospect of modelling cells {\it in silico} in unprecedented detail.  Such models will be essential to integrate the ever-increasing store of biological knowledge, and have the potential to transform medicine and biotechnology.   A key task in this emerging field of {\it systems biology\/} is to identify cellular models directly from experimental data. In applying qualitative system identification to systems biology we
focus on models of metabolism: the interaction of small molecules with enzymes (the domain of classical biochemistry). Such models are the best established in systems biology. To this end, we demonstrate that the approach scales up to identify the core of a well-known, complex biological system (glycolysis) from qualitative data. This system is modelled here by a set of $25$ qualitative relations, with
several unmeasured variables. The scale-up is achieved by augmenting the background knowledge to incorporate general chemical and biological constraints on enzymes and metabolites.

The rest of the paper is organised as follows.  In the next section we present the learning approach {\sc ILP-QSI} by means of an example: the u-tube. We also describe the details of the learning algorithm in this section. In Section 3 we apply the learning experiments to a number of other systems in the same class as the u-tube, present the results obtained, and discuss the results for all the experiments reported thus far. Section 4 extends the work from learning from qualitative data to a set of  proof-of-concept experiments to assess the ability of {\sc ILP-QSI} to  learn from quantitative data. The scalability of the {\em method} is tested in Section 5 by application to a large scale metabolic pathway: glycolysis. In Section 6 we discuss related work; and finally in Section 7 we provide a general discussion of the research and draw some general conclusions.

\section{Qualitative System Identification  Using ILP}
In order to aid understanding of the method presented in this paper we will first present a detailed description of the process as applied to an illustrative system: the u-tube. The u-tube has been chosen because it is a well understood system, and is one that has been used in the literature \cite{Muggleton90,SayKuru96}. The results emerging from this set of experiments will allow us to draw some tentative conclusions regarding qualitative systems identification.

In subsequent sections we will present the results of applying the method described in this section to further examples from the same class of system; this will enable us to generalise our tentative conclusions. We will also apply the method to a large scale biological system to demonstrate the scalability of the {\it method}.

\subsection{An Illustrative System: The U-tube}
\label{illexam}
\begin{figure}[hbt]
\begin{minipage}{92mm}
\includegraphics[scale=0.4]{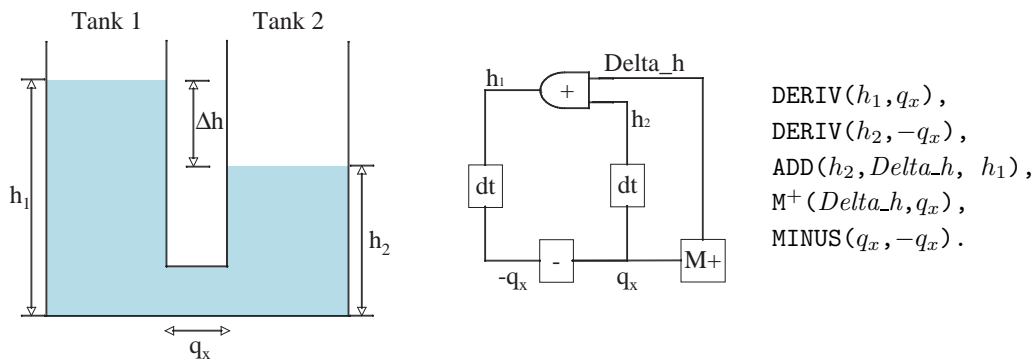}
\end{minipage} \hfill
\begin{minipage}{50mm}
\begin{small} {\tt
          DERIV($h_{1}$,$q_x$),\\
          DERIV($h_{2}$,$-q_x$),\\
          ADD($h_{2}$,{\it Delta\_h}, $h_{1}$),\\
          M$^+$({\it Delta\_h},$q_x$),\\
          MINUS($q_x$,$-q_x$).} \\
\end{small}
\end{minipage}
\caption{The u-tube: (left) physical; (middle) QSIM diagrammatic; (right) QSIM constraints. In the QSIM version of the model {\tt Delta\_h} corresponds to $\Delta h$ in the physical model. In QSIM, {\tt M$^+(\cdot,\cdot)$} is the qualitative version of a functional relation which indicates that there is a monotonically increasing relation between the two variables which are its arguments. The {\tt M$^+(\cdot,\cdot)$} constraint represents a family of functions that includes both linear and non-linear relations.}
\label{utube}
\end{figure}

The u-tube system (Fig.~\ref{utube}) is a closed system consisting of two tanks containing (or potentially containing) fluid, joined together at their base by a pipe. Assuming there is fluid in the system it passes  from one tank to the other via the pipe -- from the tank with the higher fluid level to the tank with the lower fluid level (as a function of the difference in height). If the height of fluid is the same in both tanks then the system is in equilibrium and there is no fluid flow.

The u-tube can be represented by a system of ordinary differential equations as follows: 
\begin{equation}
\left.
\begin{array}{l}
\frac{dh_1}{dt} = k \cdot (h_1 - h_2) \\
\\
\frac{dh_2}{dt} = k \cdot (h_2 - h_1) 
\label{utubeqn}
\end{array}
\right\}
\end{equation}
A qualitative model may be obtained simply by abstracting from the real numbers, which would normally be associated with  Equation~\ref{utubeqn}, into the quantity space of the signs. A common formalism used to represent qualitative models is QSIM \cite{Kuipers2}. In this representation models are conjunctions of constraints, each of which are two or three place predicates representing abstractions of  real valued arithmetic and functional operations. All variables in a model have values represented by two element vectors consisting of (in the most abstract case) the sign of both the magnitude and direction of change of the variable. In order to accommodate this restriction on the number of variables in a constraint we may rewrite  Equation~\ref{utubeqn} as follows:
\begin{equation}
\left.
\begin{array}{l}
\Delta h = (h_1 - h_2) \\
q_x = k \cdot \Delta h \\
\frac{dh_1}{dt} = q_x \\
\frac{dh_2}{dt} = -q_x
\label{rewrite}
\end{array}
\right\}
\end{equation}
where $h_1$ and $h_2$ are the height of fluid in Tank 1 and Tank 2 respectively; $\Delta h$ is the difference in the height of fluid in the tanks; and $q_x$ is the flow of fluid between the tanks. This can be converted directly to QSIM constraints as shown in Fig. \ref{utube}.

\begin{center}
\begin{table*}
 \begin{tabular}{|l|l|l|l|l|} \hline
 {\bf State} & {\bf $h_{1}$}  & {\bf $h_{2}$}  &
 {\bf $q_x$}  & {\bf $-q_x$} \\
 \hline
 1 & $<0, std>$ & $<0, std>$ & $<0, std>$ & $<0, std>$ \\
 \hline
 2 & $<0, inc>$ & $<(0, \infty), dec>$ & $<(-\infty, 0), inc>$ & $<(0, \infty),
 dec>$ \\
 \hline
 3 & $<(0, \infty), dec>$ & $<0, inc>$ & $<(0, \infty), dec>$ & $<(-\infty,
 0), inc>$ \\
 \hline
 4 & $<(0, \infty), dec>$ & $<(0, \infty), inc>$ & $<(0, \infty),
 dec>$ & $<(-\infty, 0), inc>$ \\
 \hline
 5 & $<(0, \infty),std>$ & $<(0, \infty), std>$ & $<0, std>$ & $<0,
 std>$ \\
 \hline
 6 & $<(0, \infty), inc>$ & $<(0, \infty), dec>$ & $<(-\infty, 0),
 inc>$ & $<(0, \infty), dec>$ \\
 \hline
 \end{tabular}
 \caption{\label{utubenv} The envisionment states used for the u-tube experiments. The qualitative values are in the standard form used by QSIM. Positive values for the magnitude are represented by the interval $(0, \infty)$, negative values by the interval $(- \infty, 0)$ and zero by $0$. The directions of change are self explanatory with increasing represented by $inc$, decreasing by $dec$ and steady by $std$.}
\end{table*}
 \end{center}
  \begin{figure}[btph]
\begin{center}
\includegraphics[scale=0.65]{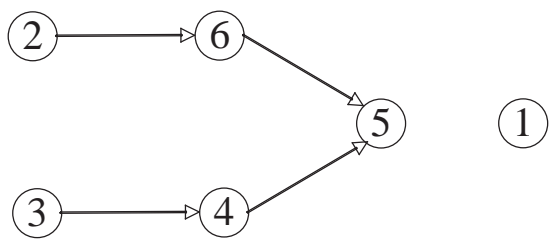}
\end{center}
\caption{\label{uenv} The u-tube envisionment graph.}
\end{figure}
Appropriate qualitative analysis of the u-tube will produce the {\it states} shown in Table~\ref{utubenv}, which are the states of the envisionment. These represent all the distinct qualitative states in which the u-tube may exist and Fig.~\ref{uenv} depicts all the possible behaviours (in terms of transitions between states)\footnote{State 1 represents the situation where there is no fluid in the system, so nothing happens and it is not interesting.}. This figure represents a {\it complete envisionment} of the system, which is the graph containing all the qualitative states of the system and all the transitions between them for a particular input value. In the case of the u-tube presented here there is no input (which is equivalent to a value of zero).  On the other hand the behaviours of a u-tube may be observed under a number of experimental (initial) conditions, with measurements being taken of the height of fluid in each tank and the flow between the tanks. These can be converted (by means of a quantitative-to-qualitative converter) into a set of qualitative observations (or states). If sufficient temporal information is available to enable the calculation of qualitative derivatives, each observation will be a tuple stating the magnitude  and direction of change of the measured variable. These observations will also contain the states in the complete envisionment of Table~\ref{utubenv} (or some subset thereof).

\begin{figure}[h]
\begin{center}
\includegraphics[scale=0.6]{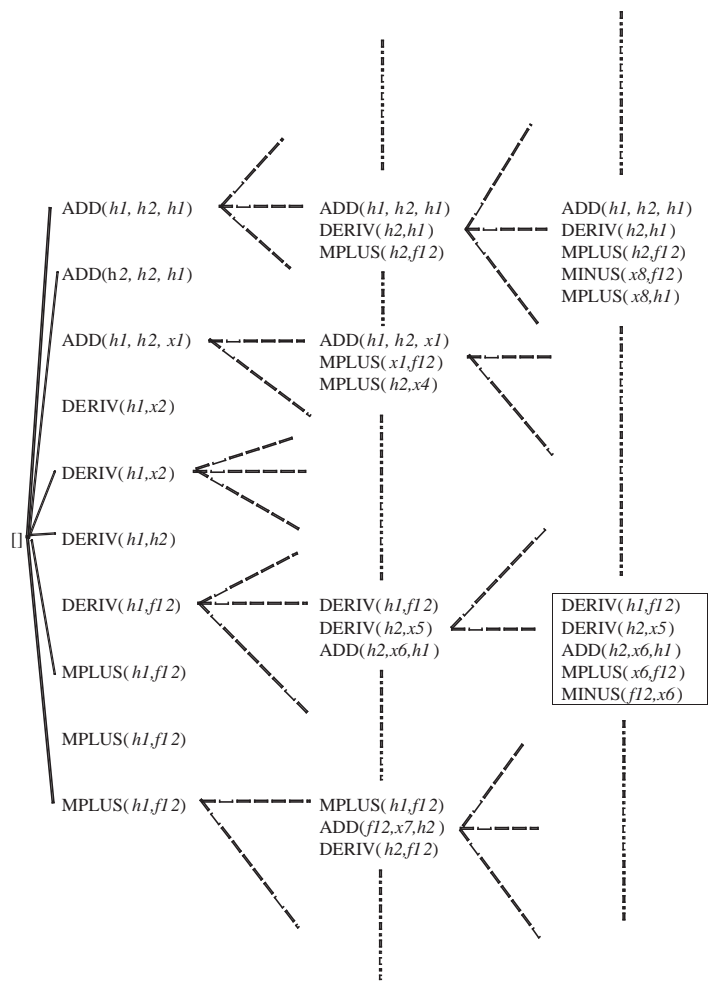}
\end{center}
\caption{\label{latt} Some portions of the u-tube lattice (with the target model in the box).}
\end{figure}

The u-tube is a member of a large class of dynamic systems which are defined by their {\em states}: state systems. In such systems the values of the variables at all future times are defined by the current state of the system regardless of how that state was achieved \cite{Healey75}. This means that for simulation, any system state can act as an initial state. In the current context it means that in order to learn the structure of such systems we need only focus on the states themselves and may ignore the transitions between states. This enables us to explore the power set of the envisionment to ascertain the conditions under which system identification is possible.
Given these qualitative observations as examples, background
knowledge consisting of constraints on models (described later) and
QSIM relations, the learning system (which we name  {\sc ILP-QSI}) performs a search for acceptable
models. To a suitable first approximation, the basic task
can be viewed as a discrete optimisation problem of finding
the lowest cost elements amongst a finite set of alternatives. That is,
given a finite discrete set of models $S$ and a
real-valued cost-function $f: {\cal S} \rightarrow \Re$, find a subset
${\cal H} \subseteq {\cal S}$ such that
${\cal H} = \{H | H \in {\cal S} ~\mathrm{and}~f(H) = {\mathrm {min}}^{}_{H_i \in {\cal S}} f(H_i)\}$.
This problem may be solved by employing a procedure that searches
through a directed acyclic graph representation of
possible models. In this representation, a pair of models are connected
in the graph if one can be transformed into another by an
operation called {\em refinement}. Fig.  \ref{latt} shows
some parts of a graph for the u-tube in which a model is
refined to another by the addition of a qualitative constraint.
An optimal search procedure (the branch-and-bound procedure)
traverses this graph in some order, at all times keeping
the cost of the best nodes so far. Whenever a node is reached where
it is certain that it and all its descendents have a cost higher than
that of the best nodes, then the node and its descendents are removed from
the search. 

%==========================================

There are a number of  features apparent in the u-tube model that are relevant to the learning method utilised in this work (and discussed in Section \ref{ILP-QSI:alg}) that will be described here since they regard general modelling issues relevant to the learning of qualitative models of dynamic systems.

\begin{figure}[btph]
\begin{center}
\includegraphics[scale=0.35]{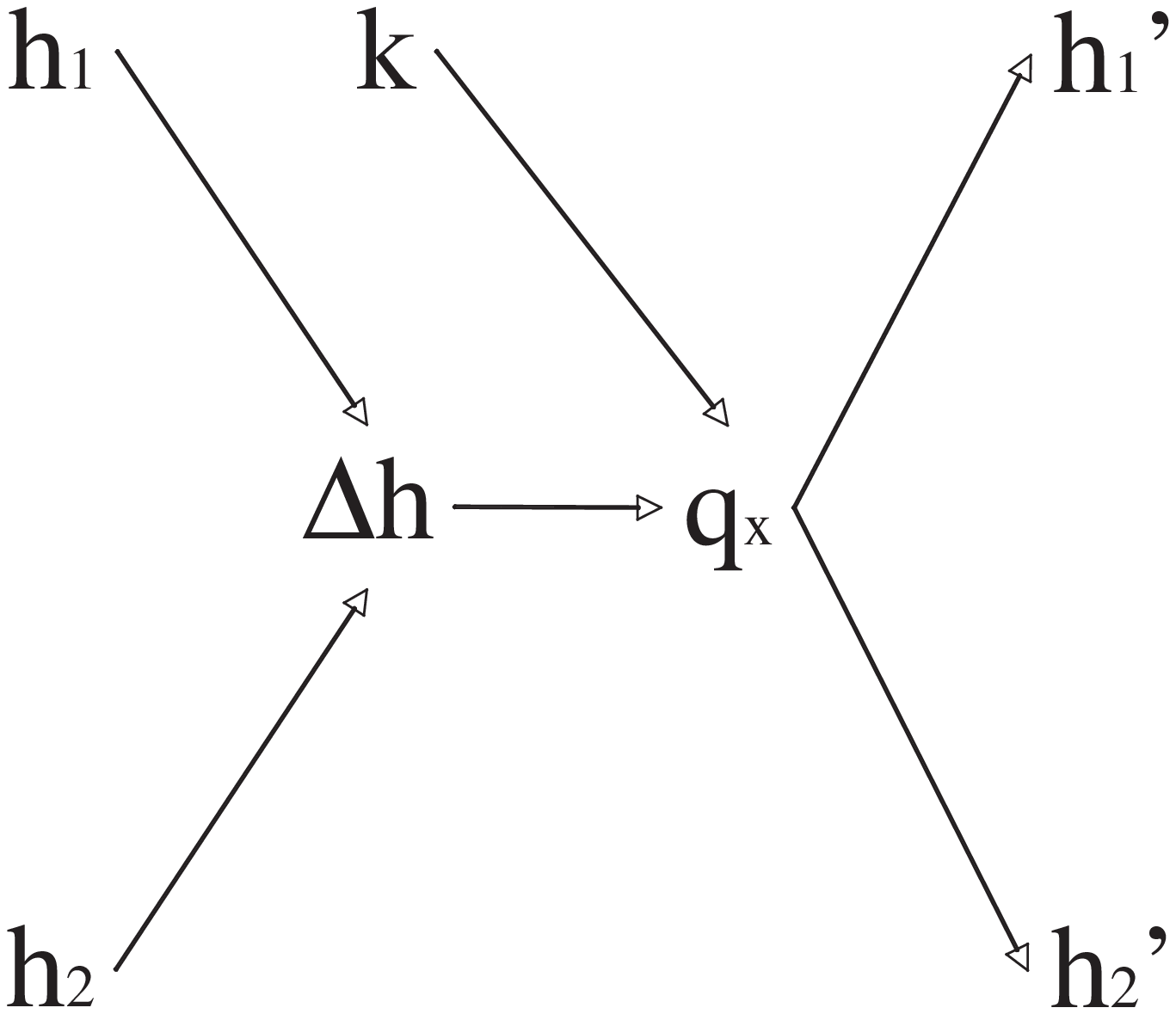}
\end{center}
\caption{\label{process}A causal ordering of the u-tube model given in Equation \ref{rewrite}.}
\end{figure}

The first thing that may be noted in this regard is that the expressions in  Equation \ref{rewrite} and the resulting qualitative constraints are \textit{ordered}; that is, given the values for the exogenous variables and the magnitude of the state variables (the height of fluid in the tanks in this case) the equations can be placed in an order such that the variables on the left hand side all may have their values calculated before they appear on the right hand side of an equation\footnote{This ordering is not required by QSIM in order to preform qualitative simulation. However, the ability to order equations in this manner can be utilised as a filter in the learning system in order to eliminate models containing algebraic loops.}. This particular form of ordering in known as \textit{causal ordering} \cite{Iwasaki86}. A causally ordered system can be depicted graphically as shown in Fig. \ref{process}.

A causally ordered model contains no \textit{algebraic loops}. In quantitative systems one tries to avoid algebraic loops because they are hard to simulate, requiring additional simultaneous equation solvers to be used.

%=============================================

A qualitative model combined with a Qualitative Reasoning (QR) inference engine will provide an envisionment of the system of interest. That is, it will generate all the qualitatively distinct states in which the system may exist. In the case of the u-tube there are six such states as given in Table \ref{utubenv}. Example behaviours resulting from these states are shown in Fig. \ref{uenv}.

%==========================================

It may be noted that the differential equation model captures the essence of the explanation given in the first paragraph of this section. It is sufficient to \textit{explain} the operation of such a system, as well as to predict the way it will behave, and it contains only those variables and constants necessary to achieve this task - i.e. the model is \textit{\textbf{parsimonious.}}\footnote{It is possible that for didactic purposes we may want to include more detail, for example a relation between the intertank flow and the pressure difference, or between the height of fluid and the pressure. There is no reason why we would expect such relations to be found; although in the context of an adequate theoretical framework into which the model fits, the model provides pointers in that direction. On the other hand, one can envisage simpler models existing which may be suitable for prediction but inadequate for the required kind of explanation. See Section~\ref{relwork} for more on this.}

Furthermore, examination of the causal diagram in Fig.~\ref{process} indicates that the causal ordering is in a particular direction -- from the magnitudes of the state variables to their derivatives. The link between the derivatives and the magnitudes of the state variables is through an integration over time. This is \textit{\textbf{integral causality}} and is the preferred kind of causality in systems engineering modelling; and simulation generally. This is because integration smooths out  noise whereas differentiation introduces it. 

All variables are either endogenous or exogenous. Exogenous variables influence the system but are not influenced by it.  \textit{\textbf{Well posed}} models do not have any \textit{flapping} variables; that is, endogenous variables that appear in only one constraint. Because QSIM includes a {\tt DERIV} constraint linking the state variables directly to their derivatives, and all the systems in which we are interested are regulatory, containing feedback paths, all endogenous variables must appear in at least two constraints. 

Well posedness and parsimony are mandatory properties of the model, the other properties are desirable but not always achievable and so may have to be relaxed. However, for all the systems examined in this paper each of these properties holds.

A final feature of the u-tube model is that it represents a single system. It is an assumption implicit in all the learning experiments described in this paper that the data measured belongs to a single coherent system. This is in keeping with general experimental approaches where it is assumed that the measurements are related in some way by being part of the same system. Of course we may get this wrong and have to relax the requirement because we discover that what we thought were related cannot actually be brought together in a single model. This generalises the requirement for parsimony in line with Einstein's adage that a model should be ``as simple as possible and no simpler''. In this case it translates to minimising the number of disjoint sub-systems identified.

\subsection{\label{qss} A Qualitative Solution Space}

In Section \ref{ILP-QSI:alg} we shall present an algorithm for automatically constructing models from data. With this method we utilise background knowledge consisting of QSIM modelling primitives combined with systems theory meta-knowledge (e.g. parsimony and causality). Later we shall also  provide an analysis of the models learned and the states utilised to learn them in order to ascertain which, if any, states are more important for successful learning. One way to facilitate this analysis is to make use of a \textit{solution space} to relate the qualitative states to the critical points of the relevant class of systems (via the isoclines of the system)\footnote{The {\em critical points} of a dynamic system are points where one or more of the derivatives of the state variables is zero. The {\em isoclines} are contours of critical points.} \cite{coghill03,coghill92}.
As stated previously, a qualitative analysis of the u-tube will generate an envisionment containing six states, as shown in Table~\ref{utubenv}, and depicted in the envisionment graph given in Fig. \ref{uenv}. 
\begin{figure}[hbt]
\begin{minipage}{80mm}
\postscript{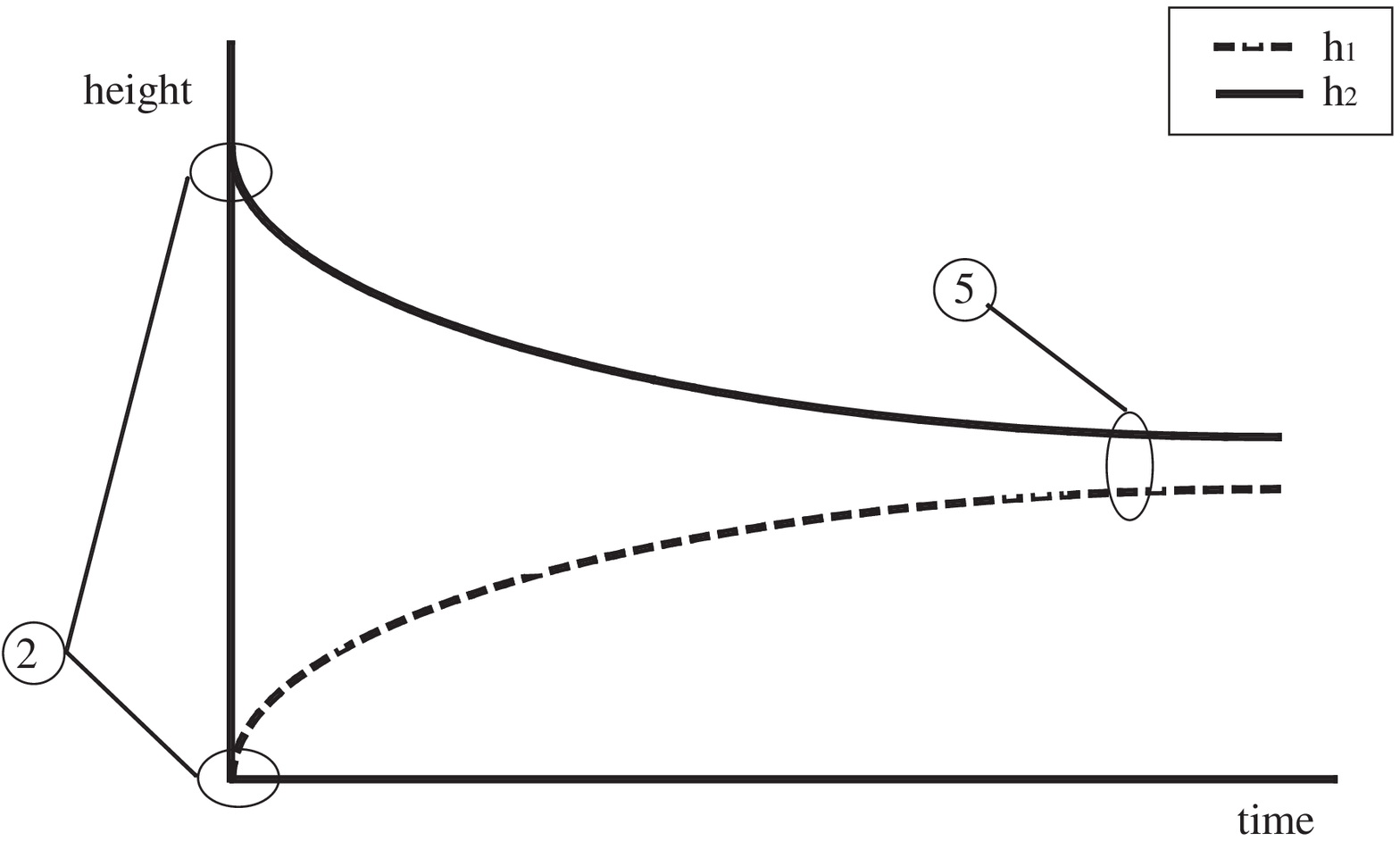}{0.98}
\end{minipage} \hfill
\begin{minipage}{80mm}
\postscript{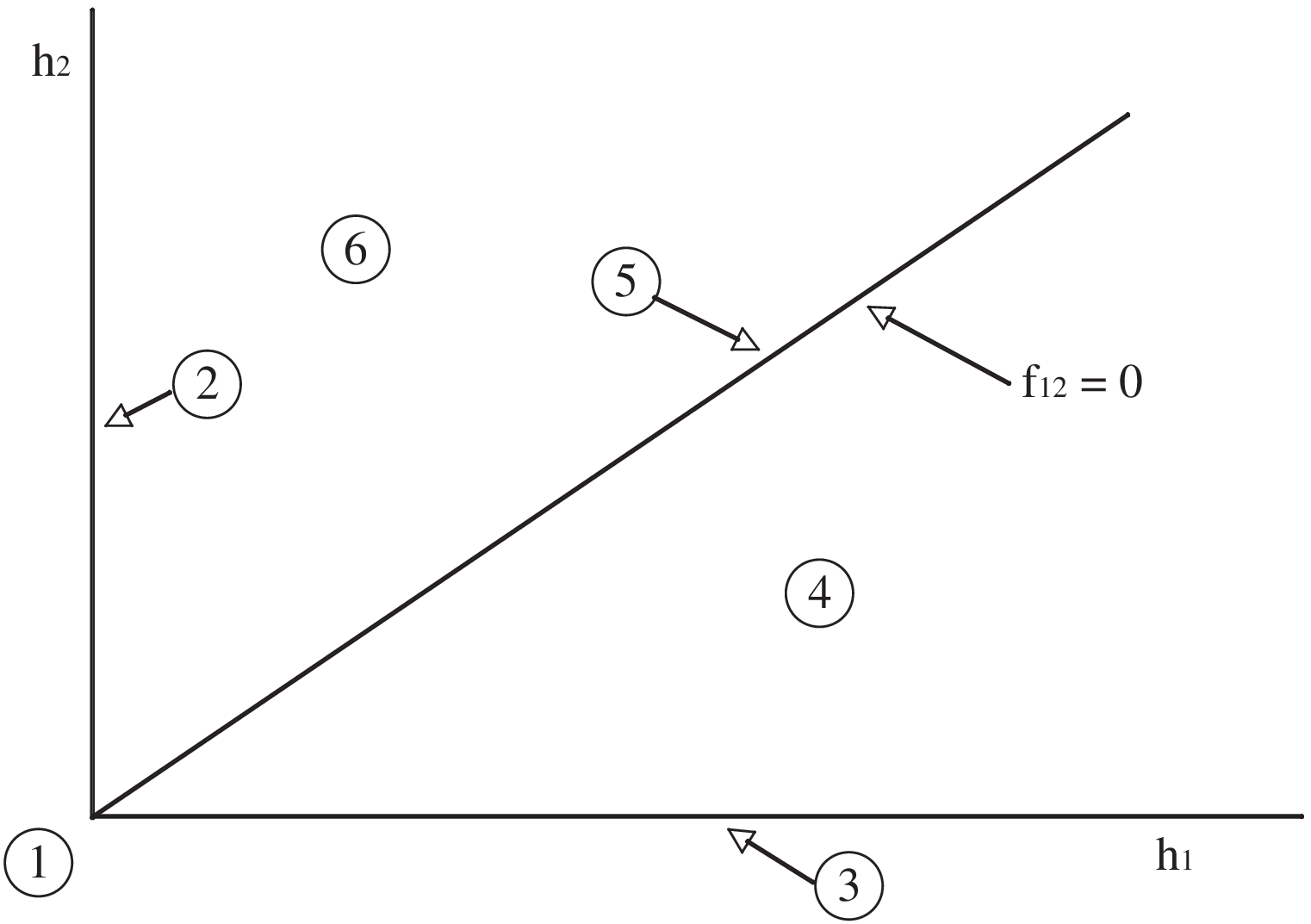}{0.98}
\end{minipage}
\caption{The qualitative states of the u-tube system presented on representative time courses (left) and on the solution space (right). The state numbers refer to the states of the u-tube described above. (State 5 represents the steady state which is strictly speaking only reached at $t = \infty$, but is in practice taken to occur when the two trajectories are ``sufficiently close'', as shown here.)}
\label{utuberes}
\end{figure}
Continuing with the u-tube; there are two ways it can behave (ignoring state 1), captured in Fig.~\ref{utuberes}. Either the head of fluid in tank 1 is greater than that in tank 2 (state 4) (in the extreme tank 1 is empty -- state 3), or the head is greater in tank 2 than tank 1 (state 6). Fig.~\ref{utuberes} (left) shows the transient behaviour for the extreme case where tank 1 is empty (state 2); it can be seen from this diagram that while the head starts in this condition its eventual end is equilibrium (state 5). In this state  Equation~\ref{utubeqn} can be rewritten as:
\begin{equation}
\left.
\begin{array}{l}
0 = k \cdot (h_1 - h_2) \\
\\
0 = k \cdot (h_2 - h_1) 
\label{utubeqn1}
\end{array}
\right\}
\end{equation}
By definition $k$ must be non-zero; so the only solution to this pair of equations is:
$$h_2 = h_1$$
This relation can be plotted on a graph as shown on the right hand side of Fig.~\ref{utuberes}. Now the qualitative states of the u-tube may be placed on this solution space graph in relation to the equilibrium line.  This representation (similar in form to a phase space diagram) is useful because it provides a global picture of the location of the qualitative states of an envisionment relative to the equilibria or critical points of the system. It has also been utilised in the construction of diagnostic expert systems \cite{warren04}. For further details of this means of analysing envisionments see the work of  \citeA{coghill03} and \citeA{coghill92}.

\begin{figure*}[htb]
{\scriptsize{
\begin{description}
\item[] $bb(i,\rho,f):$ Given an
	initial element $i$ from a discrete set $S$;
        a successor function $\rho:S \rightarrow 2^S$;
	and a cost function $f:S \rightarrow \Re$,
	return $H \subseteq S$ such that
	$H$ contains the set of cost-minimal models. That is
        for all $h_{i,j} \in H, f(h_i) = f(h_j) = f_{min}$ and
	for all $s' \in S \backslash H~f(s') > f_{min}$.
\begin{enumerate}
\item $Active := \langle (i,-\infty) \rangle$.
\item $worst := \infty$
\item $selected := \emptyset$
\item while $Active \neq \langle \rangle$
\item begin
\label{bb:startloop}
\begin{enumerate}
\item remove element $(k,cost_k)$ from $Active$
\label{bb:pop}
\item if $cost_k < worst$
\label{bb:newbest}
\item begin
\begin{enumerate}
\item $worst := cost_k$
\item $selected := \{k\}$
\item let $Prune_1 \subseteq Active$ s.t. for each
        $j \in Prune_1$, $\underline{f}(j) > worst$ where $\underline{f}(j)$ is
        the lowest cost possible from $j$ or its successors
\item remove elements of $Prune_1$ from $Active$
\end{enumerate}
\item end
\item elseif $cost_k = worst$
\label{bb:equalbest}
\begin{enumerate}
\item $selected := selected \cup \{k\}$
\end{enumerate}
\item $Branch := \rho(k)$
\item let $Prune_2 \subseteq Branch$ s.t. for each
        $j \in Prune_2$, $f_{min}(j) > best$ where $f_{min}(j)$ is
        the lowest cost possible from $j$ or its successors
\item $Bound := Branch \backslash Prune_2$
\label{bb:bound}
\item for $x \in Bound$
\begin{enumerate}
\item add $(x,f(x))$ to $Active$
\end{enumerate}
\label{bb:add}
\end {enumerate}
\item end
\item return $selected$
\end{enumerate}
\end{description}
\caption{A basic branch-and-bound algorithm. The type of
$Active$ determines specialised variants: if $Active$ is
a stack (elements are added and removed from the front) then
depth-first branch-and-bound results; if $Active$ is a queue
(elements added to the end and removed from the front) then
breadth-first branch-and-bound results; if $Active$ is a prioritised
queue then best-first branch-and-bound results.}
\label{fig:bb}
}}
\end{figure*}

\subsection{The Algorithm}
\label{ILP-QSI:alg}
The ILP learner used in this research is a multistage procedure, each of which addresses a discrete optimisation problem.
In general
terms, this is posed as follows:
given a finite discrete set $S$ and a
cost-function $f: S \rightarrow \Re$, find a subset $H \subseteq S$
such that
$H = \{s | s \in S ~\mathrm{and}~f(s) = {\mathrm {min}}^{}_{s_i \in S}
f(s_i)\}$. An optimal algorithm for solving such problems is
the ``branch-and-bound'' algorithm, shown in Fig.~\ref{fig:bb}
(the correctness, complexity and optimality properties of this
algorithm are presented in a paper by \citeR{pap:optim}). A specific variant of this
algorithm is available within the software environment
comprising {\sc Aleph} \cite{aleph}.  The modified
procedure is in Fig.~\ref{fig:alephbb}.  The principal differences
from Fig.~\ref{fig:bb} are:

\begin{enumerate}
\item The procedure is given a set of starting points $H_0$, instead
	of a single one ($i$ in Fig.~\ref{fig:bb});
\item  A limitation on the number of nodes
        explored ($n$ in Fig.~\ref{fig:alephbb});
\item The use of a boolean function
	$acceptable: {\cal H} \times {\cal B} \times {\cal E} \rightarrow \{FALSE,TRUE\}$. \\ acceptable(k,B,E) is TRUE, if and only if: (a) Hypothesis k "explains"
the examples E, given B in the usual sense understood in ILP (that is,
$B \wedge k \models E$ in the absence of noise); and (b) Hypothesis k
is consistent with any constraints $I$ contained in the background
knowledge (that is $B \wedge k \wedge I \not \models \Box$). In
practice, it is possible to merge these requirements by encoding the
requirement for entailing some or all of the examples as a constraint
in $B$;
\item Inclusion of background knowledge and examples ($B$ and $E$
        in Fig.~\ref{fig:alephbb}). These are arguments to
        both the refinement operator $\rho$ (the
	reason for this will become apparent shortly)
	and the cost function $f$.
\end{enumerate}

\begin{figure*}[htb]
{\scriptsize{
\begin{description}
\item[]$bb_A(B,E,H_0,\rho,f,n):$ Given background knowledge
	$B \in {\cal B}$; examples $E \in {\cal E}$; a non-empty set
	of initial elements $H_0$
	from a discrete set of possible hypotheses ${\cal H}$;
        a successor function
	$\rho:{\cal H} \times {\cal B} \times {\cal E} \rightarrow 2^{\cal H}$;
	a cost function $f: {\cal H} \times {\cal B} \times {\cal E} \rightarrow \Re$;
	and a maximum number of nodes $n \in \cal N$ ($n \geq 0$) to
	be explored,  return $H \subseteq {\cal H}$ such that
	$H$ contains the set of cost-minimal models of the models
	explored.	
\begin{enumerate}
\item $Active = \langle \rangle$
\item for $i \in H_0$
\begin{enumerate}
\item add $(i,-\infty)$ to  $Active$
\end{enumerate}
\item $worst := \infty$
\item $selected := \emptyset$
\item $explored := 0$
\item while ($explored < n$ and $Active \neq \langle \rangle$)
\item begin
\label{alephbb:startloop}
\begin{enumerate}
\item remove element $(k,cost_k)$ from $Active$
\item increment $explored$
\label{alephbb:pop}
\label{alephbb:cost}
\item if $acceptable(k,B,E)$
\label{alephbb:check}
\item begin
\begin{enumerate}
\item if $cost_k < worst$
\label{alephbb:newbest}
\item begin
\begin{enumerate}
\item $worst := cost$
\item $selected := \{k\}$
\item let $Prune_1 \subseteq Active$ s.t. for each
        $j \in Prune_1$, $\underline{f}(j,B,E) > worst$ where $\underline{f}(j,B,E)$ is
        the lowest cost possible from $j$ or its successors
\item remove elements of $Prune_1$ from $Active$
\end{enumerate}
\item end:
\item elseif $cost_k = worst$
\label{alephbb:equalbest}
\begin{enumerate}
\item $selected := selected \cup \{k\}$
\end{enumerate}
\end{enumerate}
\item end
\label{alephbb:endcheck}
\item $Branch := \rho(k,B,E)$
\item let $Prune_2 \subseteq Branch$ s.t. for each
        $j \in Prune_2$, $\underline{f}(j,B,E) > worst$
	where $\underline{f}(j,B,E)$ is
        the lowest cost possible from $j$ or its successors
\item $Bound := Branch \backslash Prune_2$
\label{alephbb:bound}
\item for $x \in Bound$
\begin{enumerate}
\item add $(x,f(x,B,E))$ to $Active$
\label{alephbb:add}
\end{enumerate}
\end {enumerate}
\item end
\item return $selected$
\end{enumerate}
\end{description}
\caption{A variant of the basic branch-and-bound algorithm,
implemented within the {\sc Aleph} system. Here $\cal B$ and
$\cal E$ are sets of logic programs; and $\cal N$
the set of natural numbers.} 
\label{fig:alephbb}
}}
\end{figure*}

\noindent
The following points are relevant for the implementation used here:
\begin{itemize}
\item Each qualitative model is represented as a single definite clause.
	Given a definite clause $C$, the qualitative
	constraints in the model (the size of the model)
	are obtained by counting the number of qualitative constraints in $C$.
	This will also be called the ``size of $C$''.
\item Constraints, such as the restriction to well-posed models (described
	below), are
	assumed to be encoded in the background knowledge;
\item The initial set $H_0$ in Fig.~\ref{fig:alephbb} consists
	of the empty clause denoted here as $\emptyset$. That is,
	$H_0 = \{\emptyset\}$;
\item $acceptable(C,B,E)$ is $TRUE$ for any qualitative
	model $C$ that is consistent with the constraints in $B$, given $E$.
\item $Active$ is a prioritised queue sorted by $f$;
\item The successor function used is $\rho_A$. This is defined as follows.
	Let $S$ be the size of an acceptable model and $C$ be a qualitative
	model of size $S'$ with $n = S - S'$. We assume $B$ contains
	a set of mode declarations in the form described by
	\cite{mugg:progol}. Then, given a definite
	clause $C$, obtain a definite $C' \in \rho_A(C,B,E)$ where
	$\rho_A = \rho_A^n = \langle D'|~\exists D \in \rho^{n-1}_A(C,B,E)~\mathrm{s.t.}~D' \in \rho_A^1(D,B,E)\rangle, (n \geq 2)$.
	$C' \in \rho_A^1(C,B,E)$ is obtained by adding a literal $L$ to $C$,
	such that:
	\begin{itemize}
	\item Each argument with mode $+t$ in $L$ is substituted with
		any input variable of type $t$ that appears in the positive
		literal in $C$ or with any variable of type $t$ that occurs
		in a negative literal in $C$;
	\item Each argument with mode $-t$ in $L$ is substituted with
		any variable in $C$ of type $t$ that appears before that 
		argument or by a new variable of type $t$;
	\item Each argument with mode $\#t$ in $L$ is substituted with
		a ground term of type $t$. This assumes the availability
		of a generator of elements of the Herbrand universe of terms;
		and
	\item $acceptable(C',B,E)$ is $TRUE$.
	\end{itemize}
	The following properties of $\rho_A^1$ (and, in turn of $\rho_A$)
	can be shown to hold
	\cite{riguzzi:ilp2005}:
	\begin{itemize}
	\item It is locally finite. That is, $\rho_A^1(C,B,E)$ is finite
		and computable (assuming the constraints in $B$ are
		computable);
	\item It is weakly complete. That is, any clause containing
		$n$ literals can be obtained in $n$ refinement
		steps from the empty clause;
	\item It is not proper. That is, $C'$ can be equivalent
		to $C$;
	\item It is not optimal. That is, $C'$ can be obtained
		multiply by refining different clauses.
	\end{itemize}
	In addition, it is clear by definition that
	given a qualitative model $C$, $acceptable(C',B,E)$
	is $TRUE$ for any model $C' \in \rho_A^1(C,B,E)$. In turn,
	it follows that $acceptable(C',B,E)$ is $TRUE$ for
	any $C' \in \rho_A(C,B,E)$.
\item The cost function used \cite<following>{mugg:poslearn} is $f_{Bayes}(C,B,E) = -{\mathrm P}(C|B,E)$ where
	${\mathrm P}(C|B,E)$ is the Bayesian posterior probability estimate of
	clause $C$, given background knowledge $B$ and
	positive examples $E$. Finding the model with the maximal posterior
	probability (that is, lowest cost) involves maximising the function
	\cite{mccreath:thesis}:
\[
        Q(C) = \mathrm{log}D_{\cal H}(C) + p~ \mathrm{log} {{1}\over{g(C)}}
\]
	where $D_{\cal H}$ is a prior probability measure over the
	space of possible models; $p$ is the number of positive examples
	(that is, $p = |E|$); and $g$ is the generality of a model.
	We use the approach used in 
	the ILP system C-Progol to obtain values for these two functions.
	That is, the prior probability is related to the complexity
	of models (more
	complex models are taken to be less probable, {\em a priori\/});
	and the generality of a model is estimated using the number of
	random examples entailed by the model, given the background knowledge $B$ (the
	details of this are presented by Muggleton in his paper of \citeyearR{mugg:poslearn}).
	
	We have selected this Bayesian function to score hypotheses since it
represents, to the best of our knowledge, the only one in the ILP
literature explicitly developed for the case where data consist of
positive examples only (as is the situation in this paper, where
examples are observations of system behaviour: system identification
from ``non-behaviour'' does not represent the usual understanding of
the task we are attempting here).
\end{itemize}

\noindent
It is evident that these choices make the branch-and-bound
procedure a simple ``generate-and-score'' approach. Clearly, the
approach is only scalable if the constraints encoding
well-posed models are sufficient to restrict acceptable models
to some reasonable number: we describe a set of such constraints
that are sufficient for the models examined in this paper.
In the rest of the paper, the term {\sc ILP-QSI}
will be taken to mean the {\sc Aleph} branch-and-bound algorithm
with the specific choices above.

\subsubsection{Well-posed models}
 Well-posed models were introduced in Section \ref{illexam}; in the current implementation they  are defined as satisfying at least the following syntactic constraints:

\begin{itemize}
\item[1.]{\it Size.\/} The model must be of a particular size (measured
	by the number of qualitative relations for physical models in
	Sections \ref{exp} and \ref{exprs}
	or the number of metabolites for the biological model
	in Section \ref{glycExps}). This size is pre-specified.
\item[2.]{\it Complete.\/} The model must contain all the measured
	variables.
\item[3.]{\it Determinate.\/}  The model must contain as many relations
	as variables (a basic principle of systems theory---the reader
	may recall a version from school algebra, where a system of 
	equations contains as many equations as unknowns).
\item[4.]{\it Language.\/} The number of instances of any qualitative
	relation in the model must be below some pre-specified limit. This
	kind of restriction has been studied in greater detail
	in the work of \citeA{rui:thesis}.
\end{itemize}
\noindent
and at least the following semantic constraints:

\begin{itemize}
\item[5.]{\it Sufficient.\/} The model must adequately explain the observed
	data. By ``adequate'', we intend to acknowledge here that
	due to noise in the measurements, not all
	observations may be logical consequences of the
	model\footnote{Strictly speaking, the model in conjunction with
	the background knowledge.}. The percentage of observations
	that must be explainable in this sense is a user-defined
	value. 
\item[6.]{\it Redundant.\/} The model must not contain relations that are
	redundant.
	For example, the relation {\tt ADD}$(inflow,outflow,x1)$
	is redundant if the model already has {\tt ADD}$(outflow,inflow,x1)$.
\item[7.]{\it Contradictory.\/} The model must not contain relations that are
	contradictory given other relations present in the model.
\item[8.]{\it Dimensional.\/} The model must contain relations
	that respect dimensional constraints.
	This prevents, for example, addition of relations
	like {\tt ADD}$(inflow,outflow,amount)$ that
	perform arithmetic on variables that have different units
	of measurement.
\end{itemize}

\noindent
The following additional constraints which are here incorporated in the algorithm could be
ignored (because they are preferences rather than absolute rules), but all results presented in this paper require them to be satisfied:

\begin{itemize}
\item[9.]{\it Single.\/} The model must not 
	contain two or more disjoint models.  The
	assumption is that if a set of measurements are being made within a
	particular context then the user desires a single model that includes
	those measurement variables. 
\item[10.]{\it Connected.\/}  All intermediate variables should appear
	in at least two relations.
\item[11.] {\it Causal.\/} The model must be causally
	ordered \cite{Iwasaki86} with an
	integral causality \cite{Gawthrop96}.  That is, the causality
	runs through the algebraic constraints of the model from
	the magnitudes of the state variables
	to their derivatives; and from the derivatives to the
	magnitudes through a {\tt DERIV} constraint only.
\end{itemize}

\noindent
This list is not intended to be exhaustive: we fully expect
that they would need to be augmented by other domain-specific constraints
(the biological system identification problem described in Section
\ref{glycExps} provides an instance of this). 
The advantage of using ILP is that such augmentation is possible in a
relatively straightforward manner.

\subsection{\label{exp} Experimental Investigation of Learning the U-tube System}

In this section we present a comprehensive experimental test of the learning algorithm described in the previous section.  We again focus on the u-tube to illustrate the approach and explain the results  obtained. In a subsequent section we will present the results of applying ILP-QSI to learning the structure of a number of different systems of a similar kind. The data utilised in these experiments is qualitative. It is assumed that either the measurements themselves yield qualitative values or that they are quantitative time series that have been converted to qualitative values. This latter may be necessary in situations where the quantitative time series data are not available in sufficient quantity to permit quantitative system identification to be performed.

The following is the general method applied to learning all the systems studied for this paper.

\subsubsection{Experimental Aim}

Using the u-tube system, investigate the model identification
capabilities of {\sc ILP-QSI} using qualitative data that
are subject to increasing
amounts of noise and are made increasingly sparse in order to ascertain the circumstances under which the target system may be accurately identified, as a function of the number of qualitative observations available.

\subsubsection{Materials and Method}
The model learning system ILP-QSI seeks to learn qualitative structural models from qualitative data; therefore the focus of the experiments is on learning from qualitative data. 

\paragraph*{Data}

There are no inputs (exogenous variables) to this system. The data required for learning  are combinations of the qualitative states (of which there are 6) from the envisionment shown in Table \ref{utubenv}. 

\paragraph*{Method}

There are two distinct sets of experiments reported here: those based on \textbf{noise free} data and those based on \textbf{noisy} data. The former assume that the data provided are correct and are used to test the capability of ILP-QSI in handling sparse data. The latter set of experiments captures the situation where the qualitative data  may be incorrect because of measurement errors due to noise in the original signal, or through errors introduced in a quantitative to qualitative transformation (which may occur in cases where the original data is numerical).
\\[5mm]
{\bf Noise-free data.} We use the following method for
evaluating {\sc ILP-QSI}'s system-identification performance
from noise-free data:

\noindent For the system under investigation:
\begin{enumerate}
\item Obtain the complete envisionment from specific values
        of exogeneous variables. 
        
        (In the particular case of the u-tube discussed in this section there are exogenous variables and the envisionment states are as shown in Table \ref{utubenv}, as stated above.)
\item With non-empty subsets of states in the envisionment
        as training data construct a set of models using
	{\sc ILP-QSI} and record the precision of the result.\footnote{This is the proportion
        of the models in the result that are equivalent to the correct
        model. Thus, for each training data set,
        the result returned by {\sc ILP-QSI} will have
        a precision between $0.0$ and $1.0$. The term {\em precision} as used here has the meaning usually associated with it in the Machine Learning community rather than that familiar in Qualitative Reasoning.} The number of possible non-empty sets of states for the different test scenarios for the u-tube 
is 63. ($2^N-1$, where $N$ is the number of states in the complete envisionment)
\label{method:nf_all_subsets}
\item Plot learning curves showing average precision versus
        size of training data.
\end{enumerate}

\vspace*{0.5cm}
\noindent
{\bf Noisy data.} We use the following method for evaluating {\sc ILP-QSI}'s
system-identification performance from noisy qualitative data:

\noindent For the system under investigation:
\begin{enumerate}
\item Obtain the complete envisionment from specific values
        of exogeneous variables. 
      \item Replace non-empty subsets of states in the envisionment
	with randomly generated noise states. 
	With each such combination of correct and random states,
        as training data construct a set of models using
	{\sc ILP-QSI} and record the precision of the result.\footnote{As with the non-noisy data, for each training data set,
        the result returned by {\sc ILP-QSI} will have
        a precision between $0.0$ and $1.0$.} Given a complete envisionment of $N$ states, replacing
a random subset $k > 0$ of these with random states will
result in a ``noisy'' envisionment consisting of
$N-k$ noise-free states and $k$ random states.
As with Step \ref{method:nf_all_subsets} for noise-free data,
an exhaustive replacement of all possible subsets of
the complete envisionment with random states will result in
$2^{N}-1$ noisy test sets.
\label{method:n_all_subsets}
\item Plot learning curves showing average precision versus
        size of training data.
\end{enumerate}

\subsubsection{Results}

\begin{figure}[h]
\begin{center}
\includegraphics[scale=0.6]{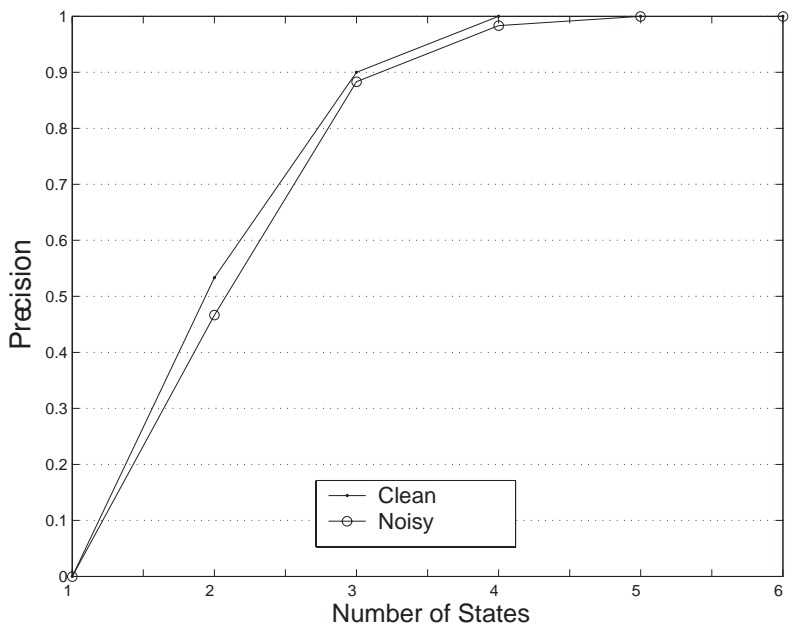}
\end{center}
\caption{\label{ut2d} Precision of models obtained for the u-tube.}
\end{figure}

The results of performing these experiments, showing the precision of learning the target model versus the number of states used (for both noise-free and noisy data) are shown in Fig. \ref{ut2d}. It is evident that for both situations precision improves with the number of states used and that the results from the experiments with noisy data have lower precision than those with the noise-free data (though the curves have the same general shape). Both these results are as one would expect.

With noise-free data we find that it was not possible to identify the target model using just one state as data. However it was possible to identify the target model using pairs of states in 53\% of cases. These states are:
\[
[2, 3], [2, 4], [2, 5], [3, 5], [3, 6], [4, 5], [4, 6], [5, 6]
\]
We refer to these as {\em Kernel sets}. For the time being we merely report this finding and delay a discussion of its significance until after reporting the results for the experiments on the other systems in the class.

\section{Experiments on Other Systems}
\label{exprs}
In this section we present the same experimental setup applied to a number of other systems: coupled tanks, cascaded tanks and a mass spring damper. These systems are representative of a class of system appearing in industrial contexts (e.g. the cascaded tanks system has been used as a model for diagnosis of an industrial Ammonia Washer system by \citeR{warren04}) as well as being useful analogs to metabolic and compartmental systems.

In each case the experimental method is identical to that utilised for the u-tube as described in Section \ref{exp}. For each system we give a description of the system and the target model, the envisionment associated with the system, a statement of the data used in the experiments, and a summary of the results obtained from the experiments.

\subsection{Experimental Aim}

For three physical systems: coupled tanks, cascaded tanks and mass-spring-damper (a well known example of a servomechanism), investigate the model identification
capabilities of {\sc ILP-QSI} using qualitative data that
are subject to increasing
amounts of noise and are made increasingly sparse.

\subsection{Materials and Method}

\paragraph*{Data}
Qualitative data available consist of the complete envisionment
arising from specific values for input variables. The precise details of the data are given with each experiment.

\paragraph*{Method}
The method used is the same as that for the u-tube and described in Section~\ref{exp}. 

\subsection{The Coupled Tanks}

\begin{figure}[hbt]
\begin{minipage}{100mm}
\postscript{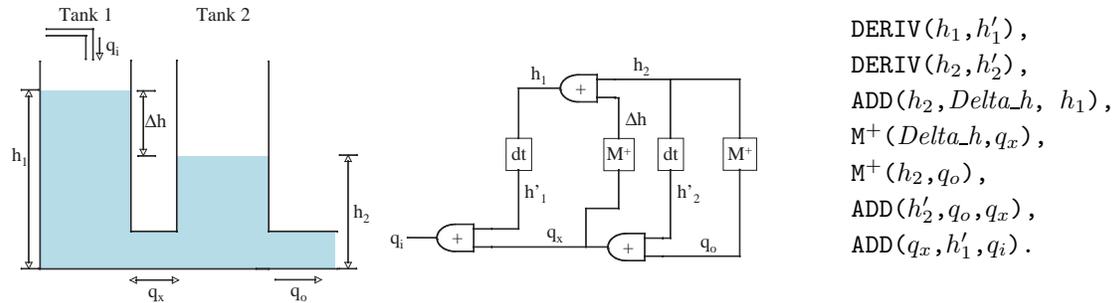}{1.0}
\end{minipage} \hfill
\begin{minipage}{40mm}
\begin{small} {\tt
          DERIV($h_{1}$,$h'_{1}$),\\
          DERIV($h_{2}$,$h'_{2}$),\\
          ADD($h_{2}$,{\it Delta\_h}, $h_{1}$),\\
          M$^{+}$({\it Delta\_h},$q_{x}$),\\
       M$^{+}$($h_{2}$,$q_{o}$),\\
       ADD($h'_{2}$,$q_{o}$,$q_{x}$),\\
       ADD($q_{x}$,$h'_{1}$,$q_{i}$).}\\
\end{small}
\end{minipage}
\caption{The coupled tanks: (left) physical; (middle) QSIM diagram; (right)
      QSIM relations.}
\label{ctanks}
\end{figure}

This is an open system consisting of two reservoirs as shown in  Fig.~\ref{ctanks}. Essentially,
it can be seen as a u-tube with an input and an output. The input, $q_i$, flows
into the top of tank $1$ and the output, $q_o$, flows out of the base of tank $2$ (see Fig.~\ref{ctanks}).  

In these experiments we assume that we can observe: $q_i$,  $q_x$, $h_1$, $h_2$, and $q_o$.  Thus system identification must discover a model with three intermediate variables, $h'_1$, 
$h'_2$ and $\Delta h$. 

\begin{figure}[h]
\begin{center}
\includegraphics[scale=0.5]{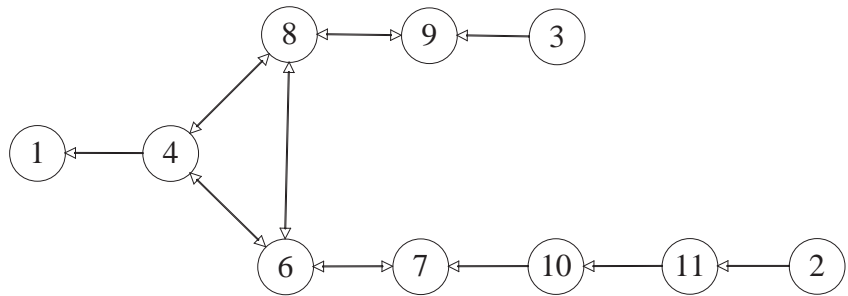}
\end{center}
\caption{\label{ctenvgraph} Coupled tanks envisionment graph.}
\end{figure}

\paragraph*{Data} There is one exogenous variable, namely the
	flow of liquid into tank 1 ($q_{i}$). In the experiments described here the
	input flow is kept at zero
	(that is, $q_{i} = \langle 0, std \rangle$), making the system
	for this particular case just moderately more complex than the u-tube.
	The complete envisionment consists of 10 states,
	as shown in Table \ref{ctanksenv} and Fig.~\ref{ctenvgraph}, which means there are 1024 experiments in this set.
	
\begin{center}
\begin{table*}
\begin{tabular}{|l|l|l|l|l|} \hline
{\bf State} & {\bf $h_1$}  & {\bf $h_2$}  &
{\bf $q_{x}$}  & {\bf $q_{o}$} \\
\hline
1 & $<0, std>$ & $<0, std>$ & $<0, std>$ & $<0, std>$ \\
\hline
2 & $<0, inc>$ & $<(0, \infty), dec>$ & $<(-\infty, 0), inc>$ & $<(0, \infty),
dec>$ \\
\hline
3 & $<(0, \infty), dec>$ & $<0, inc>$ & $<(0, \infty), dec>$ & $<0, inc>$ \\
\hline
4 & $<(0, \infty), dec>$ & $<(0, \infty), inc>$ & $<(0, \infty), dec>$ & $<(0, \infty), inc>$ \\
\hline
6 & $<(0, \infty), inc>$ & $<(0, \infty), dec>$ & $<(-\infty, 0), inc>$ & $<(0, \infty), dec>$ \\
\hline
7 & $<(0, \infty), dec>$ & $<(0, \infty), std>$ & $<(0, \infty), dec>$ & $<(0, \infty), std>$ \\
\hline
8 & $<(0, \infty), std>$ & $<(0, \infty), dec>$ & $<0, inc>$ & $<(0, \infty), dec>$ \\
\hline
9 & $<(0, \infty), dec>$ & $<(0, \infty), dec>$ & $<(0, \infty), dec>$ & $<(0, \infty), dec>$ \\
\hline
10 & $<(0, \infty), dec>$ & $<(0, \infty), dec>$ & $<(0, \infty), std>$ & $<(0, \infty), dec>$ \\
\hline
11 & $<(0, \infty), dec>$ & $<(0, \infty), dec>$ & $<(0, \infty), inc>$ & $<(0, \infty), dec>$ \\
\hline
\end{tabular}
\caption{\label{ctanksenv} The envisionment states used for the coupled tanks experiments. (The states are labeled to be in accord with those for the u-tube; since state 5 in the u-tube does not appear in the coupled tanks envisionment there is no state labeled `5' in this table.)}
\end{table*}
\end{center}

\subsubsection{Results}
\begin{figure}[h]
\begin{center}
\includegraphics[scale=0.6]{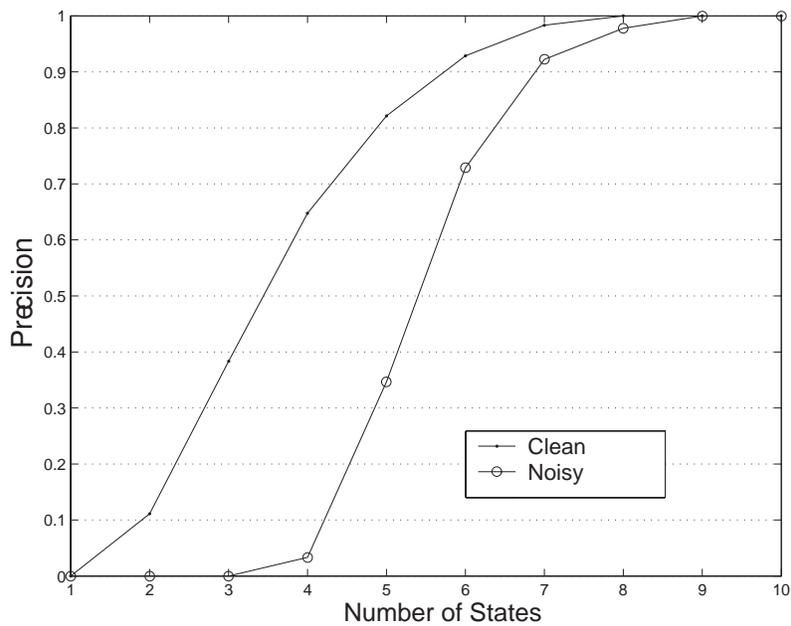}
\end{center}
\caption{\label{cprec} Coupled tanks precision graphs.}
\end{figure}

The precision graphs for the coupled tanks experiments are shown in Fig. \ref{cprec}. Here again results show the improvement in precision as the number states used increases and also the deterioration in precision when noise is added. The effect of noise is worse when fewer states are used than was the case for the u-tube, though its effect is nullified when all states are used.

For the noise free data it was again not possible to identify any models using a single datum but utilising pairs of states yielded the target model in 11\% of cases. The relevant pairs of states (kernel sets) are: 
\[
[2, 7], [3, 8], [4, 8], [6, 7], [7, 8]
\]
Whereas in the u-tube experiments all the states in which the target model was successfully learned were supersets of the pairs, in the coupled tanks case there are sets of three states (which are not supersets of the pairs listed above) that result in successful identification of the target model:
\begin{center}
$[2, 3, 9], [2, 3, 10], [2, 3, 11]  $ \\
$ [2, 4, 9], [2, 4, 10], [2, 4, 11]  $ \\
$ [3, 6, 9], [3, 6, 10], [3, 6, 11]  $ \\
$[4, 6, 9], [4, 6, 10], [4, 6, 11]$
\end{center}

\subsection{Cascaded Tanks}

This system is also an open system. However, flow through the system is always uni-directional (unlike the coupled tanks system).  In principle, the system can be broken into two sub-systems each containing one reservoir, each with their own input and output.  

\begin{figure}[hbt]
\begin{minipage}{100mm}
\postscript{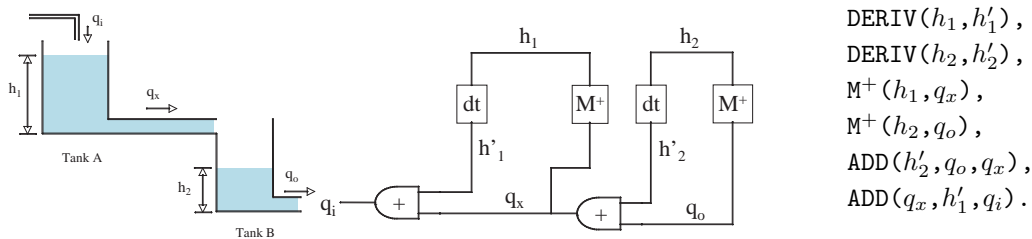}{1.0}
\end{minipage} \hfill
\begin{minipage}{40mm}
\begin{small} {\tt
       DERIV($h_{1}$,$h'_{1}$),\\
       DERIV($h_{2}$,$h'_{2}$),\\
       M$^{+}$($h_1$,$q_{x}$),\\
       M$^{+}$($h_{2}$,$q_{o}$),\\
       ADD($h'_{2}$,$q_{o}$,$q_{x}$),\\
       ADD($q_{x}$,$h'_{1}$,$q_{i}$).}\\
\end{small}
\end{minipage}
\caption{The cascaded tanks: (left) physical; (middle) QSIM diagrammatic; (right)
      QSIM relations.}
\label{castanks}
\end{figure}

An example of the system is shown in  Fig.~\ref{castanks}. Liquid flows into tank $1$, and then uni-directionally from tank $1$ into tank $2$. As is apparent from the figure, the flow is into the top of tank $1$ and out of the base
of tank $2$.  

\begin{figure}[h]
\begin{center}
\includegraphics[scale=0.75]{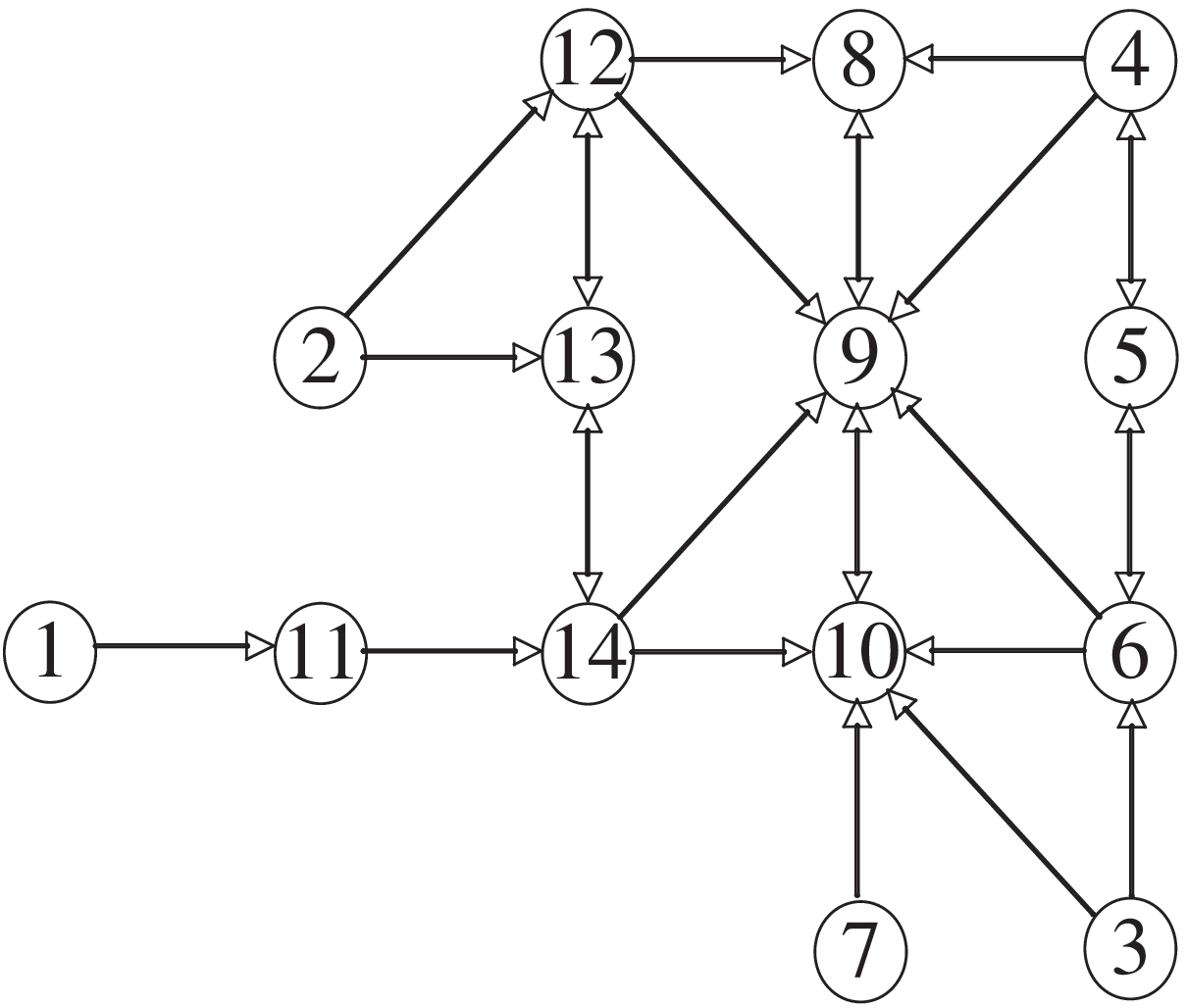}
\end{center}
\caption{\label{cascenvgraph} Cascaded tanks envisionment graph.}
\end{figure}

We assume that we can observe: $q_i$, $h_1$, $h_2$, and $q_x$.
Thus system identification must discover a model with two intermediate variables, $h'_1$ and
$h'_2$. The numbered list of states (or complete envisionment) for this
case is shown in Fig.~\ref{cascenvgraph} and Table~\ref{casenv}.

\begin{scriptsize}
\begin{center}
\begin{table*}
\begin{tabular}{|l|l|l|l|l|} \hline
{\bf State} & {\bf $h_1$}  & {\bf $h_2$}  &
{\bf $q_{x}$}  & {\bf $q_{o}$} \\
\hline
1 & $<0, inc>$ & $<0, std>$ & $<0, inc>$ & $<0, std>$ \\
\hline
2 & $<0, inc>$ & $<(0, \infty), dec>$ & $<0, inc>$ & $<(0, \infty),
dec>$ \\
\hline
 3 & $<(0, \infty), dec>$ & $<0, inc>$ & $<(0, \infty), dec>$ & $<0, inc>$ \\
\hline
4 & $<(0, \infty), dec>$ & $<(0, \infty), dec>$ & $<(0, \infty),
dec>$ & $<(0, \infty), dec>$ \\
\hline
5 & $<(0, \infty), dec>$ & $<(0, \infty),  std>$ & $<(0, \infty),
dec>$ & $<(0, \infty), std>$ \\
\hline
6 & $<(0, \infty), dec>$ & $<(0, \infty), inc>$ & $<(0, \infty),
dec>$ & $<(0, \infty), inc>$ \\
\hline
 7 & $<(0, \infty), std>$ & $<0, inc>$ & $<(0, \infty),
 std>$ & $<0, inc>$ \\
 \hline
 8 & $<(0, \infty), std>$ & $<(0, \infty), dec>$ & $<(0, \infty),
 std>$ & $<(0, \infty), dec>$ \\
 \hline
 9 & $<(0, \infty), std>$ & $<(0, \infty), std>$ & $<(0, \infty), std>$ &
 $<(0, \infty), std>$ \\
 \hline
 10 & $<(0, \infty), std>$ & $<(0, \infty), inc>$ & $<(0, \infty), std>$ &
 $<(0, \infty), inc>$ \\
 \hline
 11 & $<(0, \infty), inc>$ & $<0, inc>$ & $<(0, \infty),
 inc>$ & $<0, inc>$ \\
 \hline
 12 & $<(0, \infty), inc>$ & $<(0, \infty), dec>$ & $<(0, \infty),
 inc>$ & $<(0, \infty), dec>$ \\
 \hline
 13 & $<(0, \infty), inc>$ & $<(0, \infty), std>$ & $<(0, \infty),
 inc>$ & $<(0, \infty), std>$ \\
 \hline
 14 & $<(0, \infty), inc>$ & $<(0, \infty), inc>$ & $<(0, \infty),
 inc>$ & $<(0, \infty), inc>$ \\
 \hline
 \end{tabular}
\caption{ \label{casenv} The envisionment states used for the cascaded tanks experiments.}
\end{table*}
 \end{center}
\end{scriptsize}

\paragraph*{Data} There is one exogenous variable, namely the flow of liquid into tank $1$ ($q_{i}$). We increase the complexity by allowing a steady positive input flow (that is, $q_{i} = \langle (0, \infty), std \rangle$). The complete envisionment consists of 14 states, as shown in Fig.~\ref{cascenvgraph} and Table~\ref{casenv} which means 16,383 experiments are required .

\subsubsection{\label{cascResult} Results}

\begin{figure}[h]
\begin{center}
\includegraphics[scale=0.6]{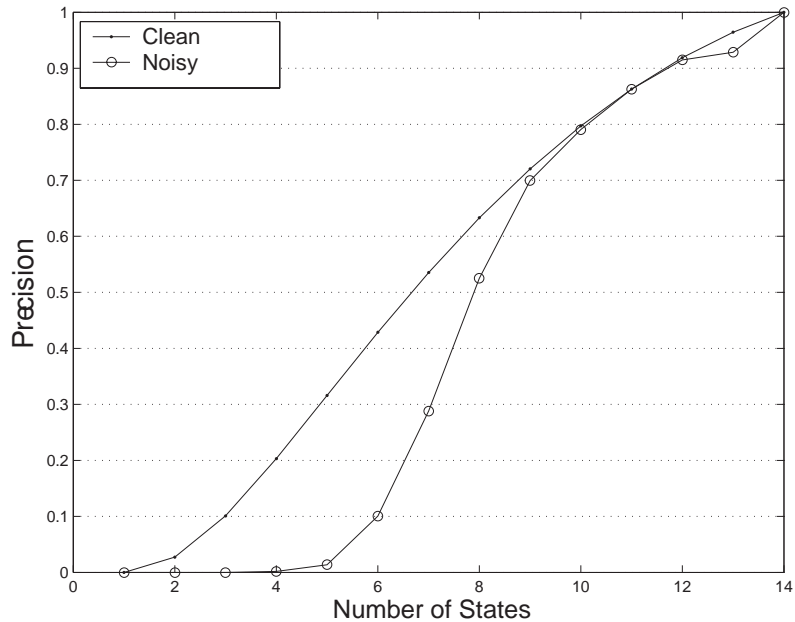}
\end{center}
\caption{\label{cs1} Cascaded tanks precision graph.}
\end{figure}

The precision graphs for the cascaded tanks are shown in Fig.~\ref{cs1}. The graphs are similar in shape to the coupled systems, but showing generally lower precision with noisy-data. Further examination shows that we are unable to identify the target model from fewer than three states.  The  subset triples (which form the kernel sets in this case) from which the target model was identified are:
\[
[1,3,4], [1,3,5], [1,3,8], [1,3,9], [1,7,4], [1,7,5], [1,7,8], [1,7,9]
\]

\subsection{Mass-Spring-Damper}

\begin{figure}[htb]
\begin{minipage}{85mm}
\postscript{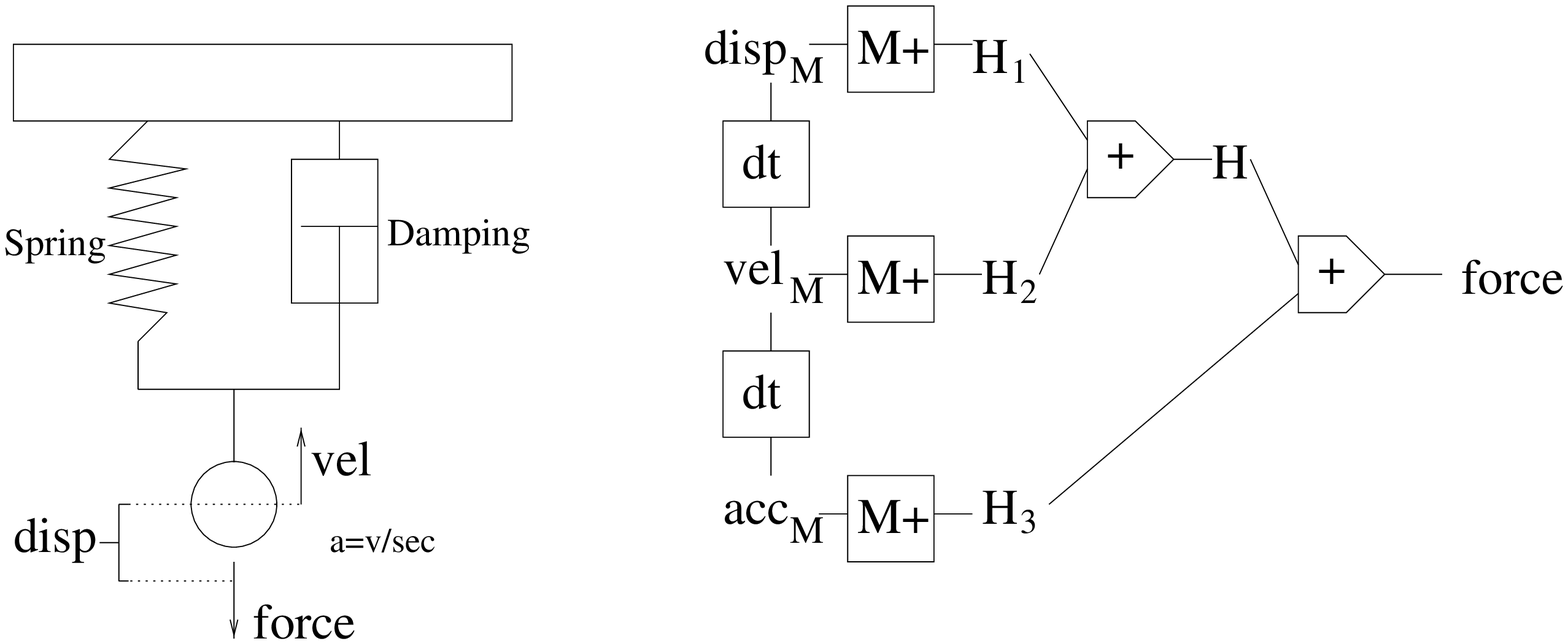}{1.0}
\end{minipage} \hfill ~ \hfill
\begin{minipage}{60mm}
\begin{small} {\tt
         DERIV($disp_M$,$vel_M$),\\
         DERIV($vel_M$,$acc_M$),\\
         M$^{+}$($disp_M$,$H_1$),\\
 	M$^{+}$($vel_M$,$H_2$),\\
 	M$^{+}$($acc_M$,$H_3$),\\
 	ADD($H_1$,$H_2$,$H_4$),\\
 	ADD($H_3$,$H_4$,{\it force}).}\\
\end{small}
\end{minipage}
\caption{The spring system (a) physical; (b) QSIM diagrammatic; (c) 
	QSIM relations}
\label{spring}
\end{figure}

\begin{figure}[h]
\begin{center}
\includegraphics[scale=0.4]{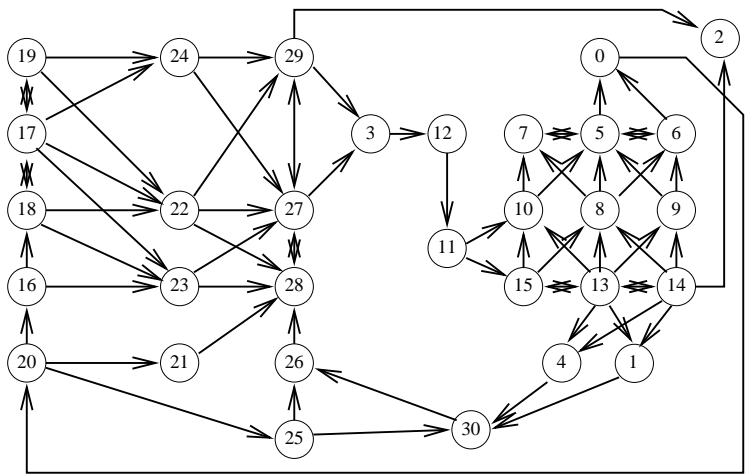}
\end{center}
\caption{\label{springenvgraph} Mass-spring-damper envisionment graph}
\end{figure}

\begin{figure}[h]
\begin{center}
\includegraphics[scale=0.6]{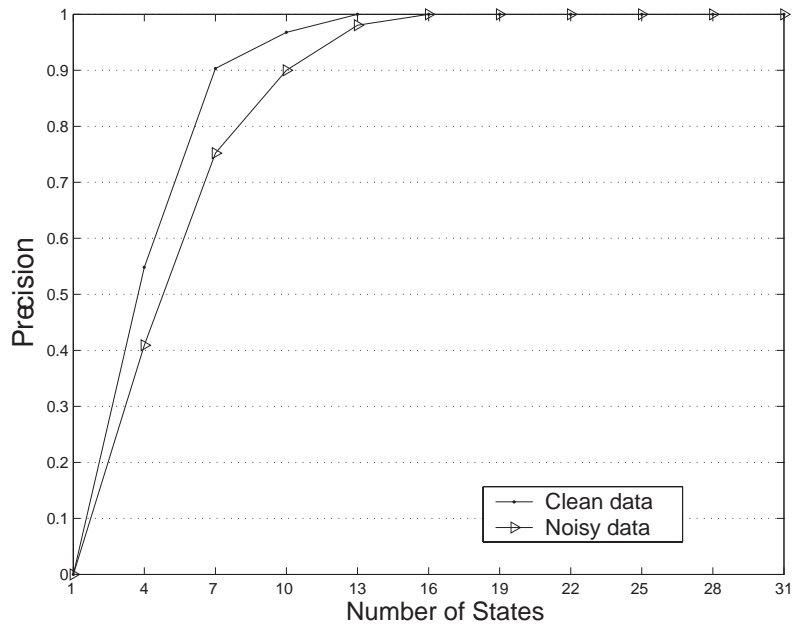}
\end{center}
\caption{\label{gr4} Mass-spring-damper precision graph}
\end{figure}

The final physical system considered is an abstraction of a wide variety of servomechanisms with a displacing force. An example of the system is shown in Fig.~\ref{spring}. In this situation, a mass is held in equilibrium between two forces.  If the equilibrium is disturbed, oscillatory behaviour is observed.  The motion of the mass is damped so that the oscillations do not continue indefinitely, and will eventually return to the original equilibrium position.  If an external force is applied (for example pulling the mass down) its final resting place will be displaced from the natural equilibrium point (see Fig.~\ref{spring}).  The mass $M$ has displacement $disp_{M}$ from its rest position, and at any time, $t$, it is moving with velocity $vel_{M}$ and accelerating at rate $acc_{M}$.
We assume that we can observe the variables: $disp_M$, $vel_M$, $acc_M$, and $force$.  Qualitative system identification must now find a model with four intermediate variables, $H_1$, $H_2$, $H_3$ and $H_4$; as well as a intermediate relation ADD($H_1$,$H_2$,$H_4$), between three of these variables.
The input force, {\it force}, is exogenous. In the experiments here, we only consider the
case where there is a steady force being applied to the system (that is, {\it Force}$_{A} =$ $\langle(0, \infty), std \rangle$). The complete envisionment for this case is shown in Fig.~\ref{springenvgraph}, where the equilibrium point is represented by state 2. 
The precision graphs are shown in Fig.~\ref{gr4}.  For this system the envisionment contains 31 states, which makes exhaustive testing unrealistic. Instead sets of clean and noisy states were randomly selected from the space of possible experiments. Nonetheless it can be observed that the average precision graphs are in-line with those obtained for the tanks experiments. However, the actual precision values suggest that both sparse data and noise have less of an effect here than the other systems. This may be due to the tight relationship between the two derivatives in the spring model, making the system extremely constrained.

\subsection{Discussion of Results}

An inspection of the experimental results reveals an expected pattern: in all cases the precision curves (for both noisy and noise free experiments) have the same general shape. Experiments which utilise fewer states identify the target model less often than when a greater number of states are used. However, a closer examination of the results reveals that even when few states are used (pairs or triples) the target model may be consistently found when particular combinations of states are used. In order to understand why this is so requires us to look at the solution spaces for the systems concerned.\footnote{We do not discuss the spring system here because of its complexity.}

We will examine the u-tube and coupled tanks together because they are very closely related systems and both had zero input. The cascaded tanks system is slightly different and had a non-zero input and so will be discussed later in the section.

\subsubsection{The U-tube and Coupled Tanks}

The bare results, while interesting, do not give any indication of why the particular pairs or triples highlighted should precisely identify the target model. In order to ascertain `why?' we must examine the envisionment states given in Tables \ref{utubenv} and \ref{ctanksenv}, from these we can itemise the relevant features of the sets of states as follows:

\begin{itemize}

\item For both the u-tube and coupled tanks there is at least one critical point in each pair.

\item For both these systems each pair of states contains one state for each branch in the envisionment graph (Fig.~\ref{uenv} \& Fig.~\ref{ctenvgraph}); and of these at least one is at the extreme of its branch. That is, states where one tank is either empty or the state immediately succeeding this, and the other tank is relatively full so that that the derivatives for that height in each tank have opposite signs.

\item For both systems all supersets of these minimal sets will precisely learn the target model.

\end{itemize}

These observations lead us to suggest that for coupled systems the ability of the learning system to identify the structure of a model is dependent on the data used including the critical points and on having data that covers all the different types of starting point that the system behaviours can have. This is in keeping with what systems theory would lead us to expect \cite{Gawthrop96}.

In order properly to appreciate what is indicated by these kernel sets  and the relation of the systems to each other we need to  look at the solution spaces \cite{coghill92,coghill03} for the two systems. These are shown in Fig. \ref{tss} (and their derivation is similar to that given in Section~\ref{qss} and detailed in Appendix \ref{solutionSpace}). From these we can get a clear picture of where the kernel pairs and triples lie with respect to the critical points of the system.

\begin{figure}[hbtp]
\begin{center}
\includegraphics*[scale=0.55]{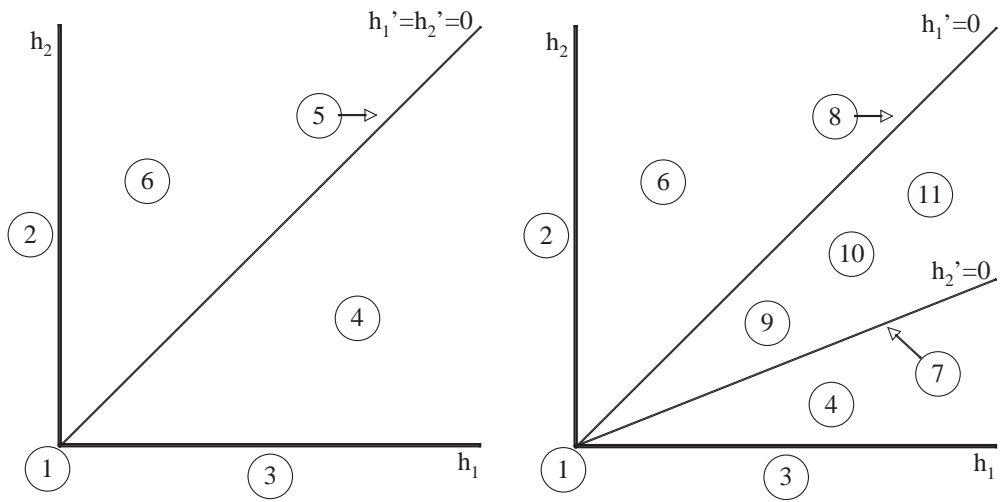}
\end{center}
\caption{\label{tss} The solution spaces for the u-tube and coupled tanks systems.}
\end{figure} 

From the system diagrams provided in Fig.~\ref{utube} and Fig.~\ref{ctanks} it can be seen that the u-tube and coupled tanks systems differ only in the fact that the coupled tanks has an outlet orifice, whereas the u-tube does not. This accounts for the major difference in their solution spaces; namely that the coupled tanks has two critical points (states 7 and 9) whereas the u-tube has only one (state 5) -- which is actually the steady state. This gives rise to the additional states: 9, 10 and 11 which lie between the critical points.\footnote{There are three states here because they differ only in the magnitude of the $q_x$ or $qdir$ of $h'_1$ and $h'_2$, neither of which appear explicitly in the solution space. Readers may convince themselves of this by comparing Table \ref{utubenv} with the envisionment in Table \ref{ctanksenv}.} It can be observed that as the outlet orifice from tank 2 in the coupled tanks system decreases in size the space between the isoclines in the solution space will become narrower until it disappears when the orifice closes. This can be seen more formally by comparing equations \ref{ute} and \ref{ce4} in Appendix \ref{solutionSpace}. There it is clear that as $k_2$ approaches zero, equation \ref{ute} approximates equation \ref{ce4} (and when $k_2 = 0$ the two equations are the same).

If we now look again at the sets of pairs we can observe that they are related in ways that reflect the relationship between the two coupled systems. Firstly, looking at the pairs. For the u-tube there are 4 pairs which include the critical point (steady state), state 5: [2, 5], [3, 5], [4, 5], and [5, 6]. Now noting from the discussion above that state 5 in the u-tube relates to either of states 7 or 8 in the coupled tanks then we find that the analogous pairs exist in the kernel set for the coupled tanks: [2, 7], [3,8], [4, 8], and [6, 7]. This leaves one pair from the coupled tanks pairs unaccounted for: [7, 8]. However, this is no surprise since that pair is taken to map to state 5 in the u-tube; and it is the consistent finding that no singleton state is sufficient to learn a model of the system.

There are still 4 pairs in the u-tube experiments from which we are able to learn reliably the target model that do not have a corresponding coupled tanks pair. These are: [2, 3], [2, 4], [3, 6], and [4, 6]. A comparison with the triples for the learning of the coupled tanks model reveals that these states  are the pairs which are conjoined with either state 9, 10 or 11 to make up the triples. The inclusion of these states warrants further explanation since they are the states which distinguish the closed u-tube from the open coupled tanks. In all three of these states the state variables both have the value $\langle(0, \infty), dec)\rangle$; a situation that cannot occur in the u-tube. Combining this with the fact that the four pairs listed above do not contain a critical point and are qualitatively identical in both systems leads one to the conclusion that the additional information contained in these triple kernel sets enables one to distinguish between the u-tube and coupled tanks in such a case.

These results extend, strengthen and deepen those reported in \shortciteA{coghill04} and \shortciteA{garrett07}.

\subsubsection{The Cascaded Tanks}

The cascaded tanks system is asymmetrical with the flow only being possible in one direction. The fact that the input is a positive steady flow makes the setup marginally more complex in that regard than for the coupled systems, where there was no input flow.

\begin{figure}[hbtp]
\begin{center}
\includegraphics[scale=0.75]{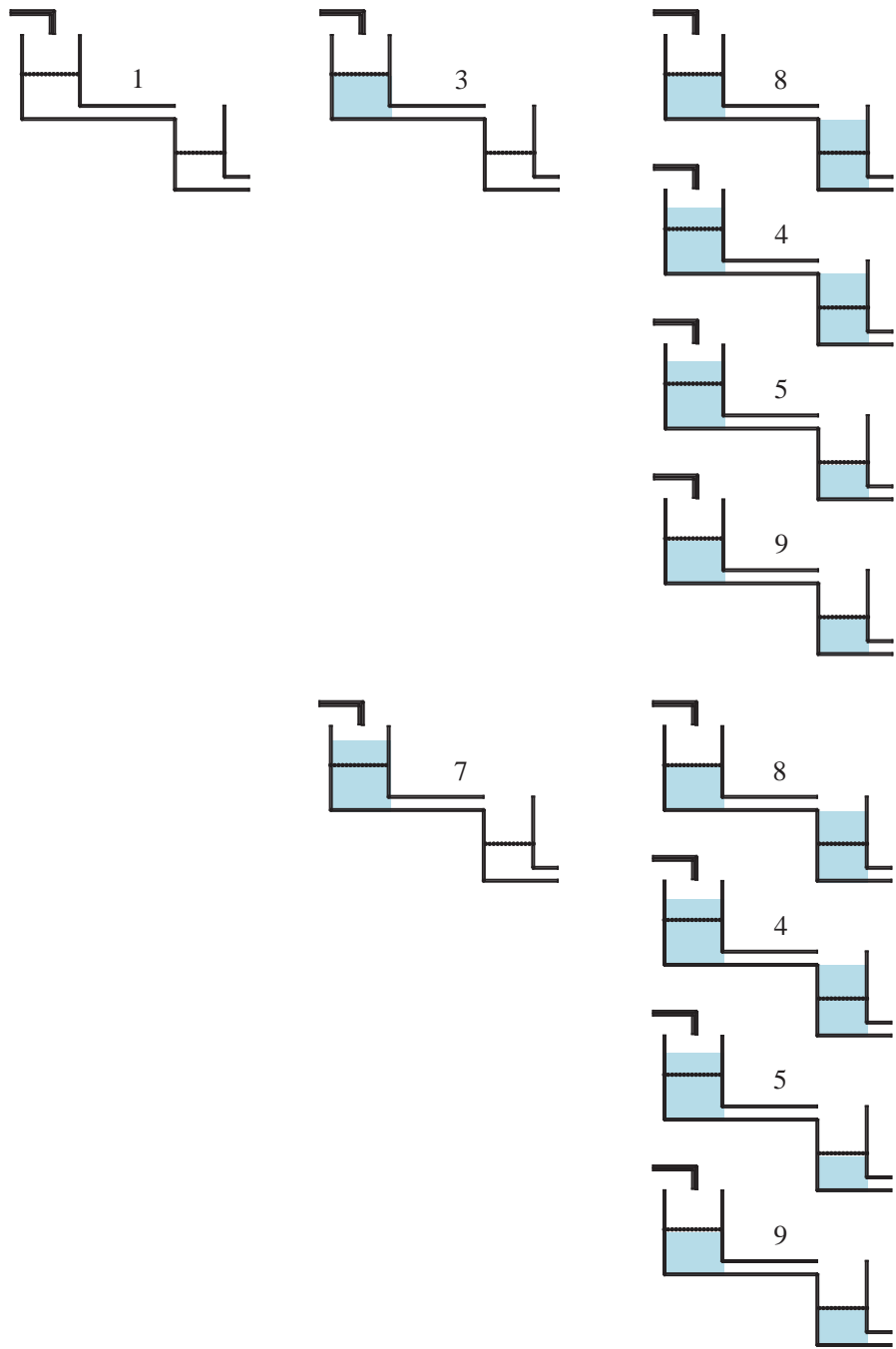}
\end{center}
\caption{\label{casres} A schematic representation of the triples of states from which the target model for the cascaded tanks systems is learned.}
\end{figure}

The kernel sets from which a model of this system may be learned (presented in Section \ref{cascResult} are depicted schematically in Fig. \ref{casres} in order to explain the results obtained. If we look at the first and middle columns of this diagram and ignore, for the time being, the downstream tank, we can see that what is represented are two pairs of states: the tank empty combined with the tank being at steady state, or the tank empty combined with the state where the amount of fluid in the tank is greater than steady state. We have confirmed experimentally that these are kernel sets from which a single tank model can be learned.

If we now ignore the upstream tank (apart from its outflow) and examine the middle and third columns of the diagram we can see that these divide into two groups according to whether the input to the downstream tank is steady or decaying (positive and decreasing). For each of these there are two pairs of states, which are the same as for the upstream tank:  the tank empty combined with the tank being at steady state, or the tank empty combined with the state where the amount of fluid in the tank is greater than steady state. In the case of this tank it can be seen that the cross product of states appear in the kernel sets because each case represents a valid possible situation.

These results lead to two major conclusions with regard to the cascaded tanks system:

\begin{enumerate}

\item ILP-QSI effectively identifies the individual components of the cascade and combines them through the cascade point.

\item The situation with the downstream tank, where the input was either a steady flow or a decreasing flow, indicates that utilising a variety of inputs can aid in the identification process.

\end{enumerate}

The former conclusion may serve as a pointer to the possibility of incremental learning of cascaded systems.

\section{Experiments with Quantitative Data}

This part of the experimental testing of our system is a `` proof-of-concept'' test.
As has been stated our system has been designed to learn qualitative models from qualitative data. As such it is assumed that the conversion of any quantitative data has already been performed, not least because the needed qualitative data analysis would require another research project and is beyond the scope of this paper. However, in order to test the usability of our system with quantitative data and to test its ability to go through the whole process from receiving the data to producing the model, we have implemented a rudimentary  data analysis package to facilitate this. Of course this is not exhaustive, but it will permit us to test the results produced via such a process for consistency with those produced from the experiments with qualitative data.

\subsection{Experimental Aim}

Using the four physical systems, investigate the model identification
capabilities of {\sc ILP-QSI} using numeric traces of system behaviour
that are subject to increasing amounts of noise.

\subsubsection{Quantitative to Qualitative Conversion}
\label{appb}

Before proceeding to describe the experiments carried out, we present the method used to convert numerical data into the qualitative form required by {\sc ILP-QSI} because this is utilised in each set of experiments.

We have adopted a straightforward and simple approach
to performing the conversion. For a quantitative variable $x$, values at $N$
real-valued time series steps were numerically
differentiated by means of a central difference approach
\cite{Shoup79} such that,

\[ \left. \begin{array}
{lll} \frac{dx_{i}}{dt} & = & \frac{(x_i - x_{i-1}) + (x_{i+1} - x_i)}{2} \\[10pt]
\frac{d^2x_{i}}{dt^2} & = & (x_i - x_{i-1}) - (x_{i+1} - x_i) \\ \end{array}
\right\}  i = 2 \cdots N - 1 \]

A quantitative variable $x$  is converted into a qualitative
variable $q$ = $\langle qmag, qdir \rangle$, where
$qmag \in$ \{(-$\infty$,0), 0, (0,$\infty$)\} is generated from $x$,
and $qdir \in \{dec, std, inc\}$ is generated from $dx/dt$.  The
qualitative derivative of $q$, $\dot{q}$, is obtained in a similar manner but
is generated from from $dx/dt$ and $d^2x/dt^2$ respectively.  

The data are typically noisy---either inherently,
or because of the process of differentiation---and
we perform some simple smoothing of
the first and second derivatives using a Blackman filter
\cite<a relative of the moving average filter -- see>{Blackman58}.
In each case, the filter is actually applied to the result of 
a Fast Fourier Transform (FFT) and the result obtained by taking the real
part of the inverse FFT. We note here that this form of smoothing is 
appropriate only when a sufficient number of time
steps are present.

Having obtained a (smoothed) numerical value $x_i$
for variable $x$ at instant $i$, its qualitative magnitude
${qmag}(x_i)$ is, in principle, simply obtained by the following:

\[
qmag(x_i) = \left \{ \begin{array}{ll}
			(-\infty,0) & \mbox{if $x_i < 0$} \\
			0 & \mbox{if $x_i = 0$} \\
			(0,+\infty) & \mbox{otherwise}
			\end{array}
		\right.
\]

In practice, since floating-point values are unlikely
to be exactly zero, we have found it advantageous
to re-apply the filtering process to data straddling zero to eliminate small fluctuations
around this value.
Despite these measures, 
in addition to generating correct qualitative states (true positives)
the conversion can produce errors: states generated may not correspond to
true states (false positives); and some true states may not be generated
(false negatives). Fig. ~\ref{fig:noiseTable} shows an example of
this (the problem is, of course, exacerbated further if the original
quantitative data are noisy). The reason for these imperfect results from noise free quantitative data are twofold: one is the smoothing process on small fluctuations around zero; but the main reason is that, as discussed above, creating a full qualitative state involves numerical differentiation which introduces noise into the data for the derivatives that affects the ability of the process to convert from quantitative to qualitative with absolute accuracy.

\begin{figure}[htb]
\begin{center} \vspace{1ex}
{\small{
\begin{tabular}{|cccccc|} \hline \hline
System & True & Generated & True & False & False \\ 
       & States& States & Positives & Positives & Negatives \\ \hline
u-tube   & 6  & 6 & 4 & 2 & 0 \\ 
Coupled  & 10 & 14 & 6 & 8 & 0 \\ 
Cascaded & 14 & 8 & 5 & 3 & 6 \\
Spring   & 33 & 33 & 20 & 13 & 0 \\ \hline
\end{tabular}
}}
\end{center}
\caption{An example of the errors resulting from generating
	qualitative states from traces of
	system behaviour.  Here, the traces were generated by
	the following initial conditions:
	$h_1 = 2.0, h_2 = 0.0$ for all
	three tank systems; and $disp_M = 2.0$, $vel_M = 0.0$
	for the spring.}
\label{fig:noiseTable}
\end{figure}

\subsection{Materials and Method}
Numerical simulations of the four physical systems were
constructed using the same general relations as the qualitative models. Once again the experiments were carried out utilising both noise free and noisy data, as described in the rest of this section.

\subsubsection{Data}

\begin{figure}
\begin{center}
\begin{minipage}{65mm}
	\postscript{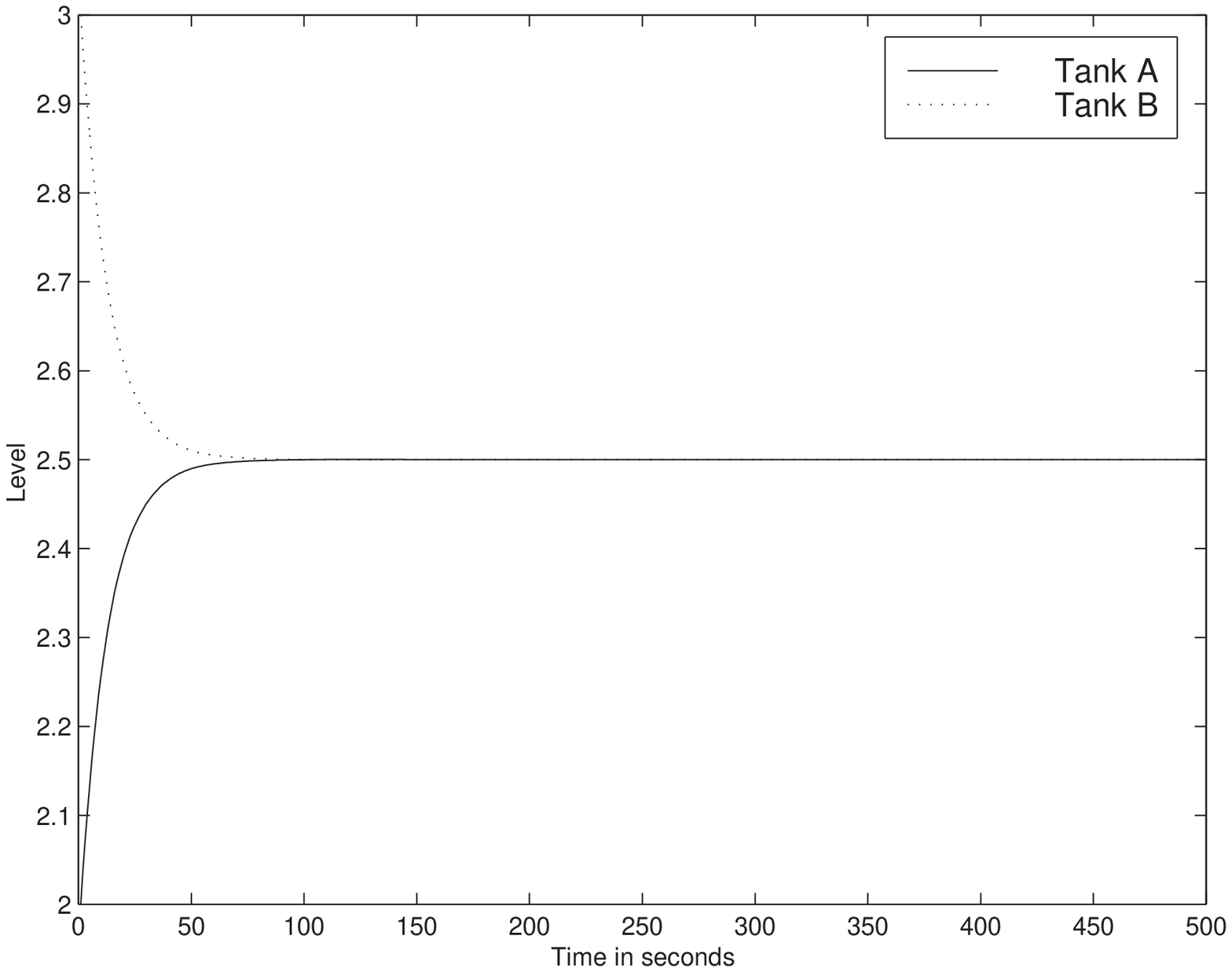}{1.0}
\end{minipage} \hfill
\begin{minipage}{65mm}
	\postscript{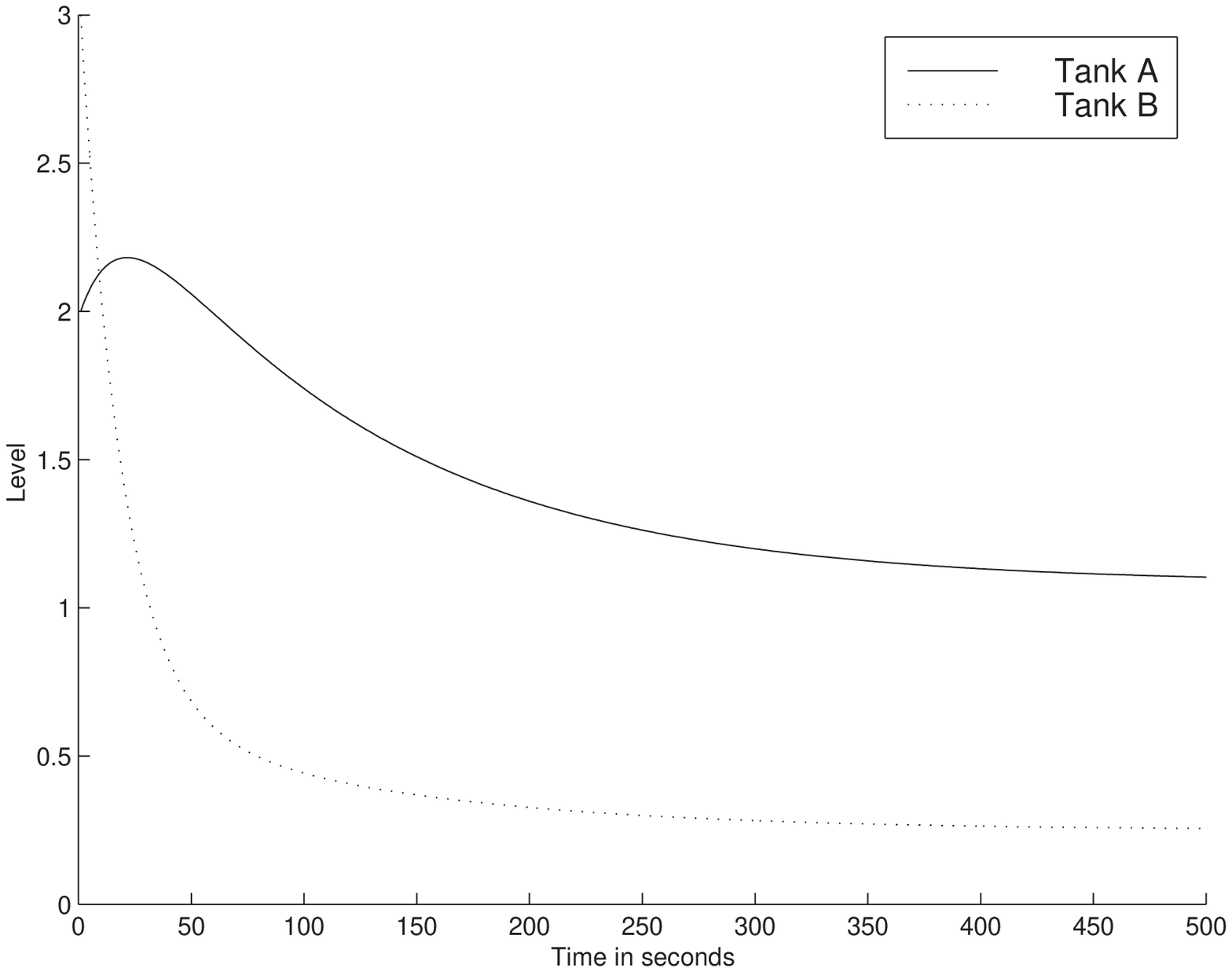}{1.0}
\end{minipage} \\  ~  \\ ~ \\
\begin{minipage}{65mm}
	\postscript{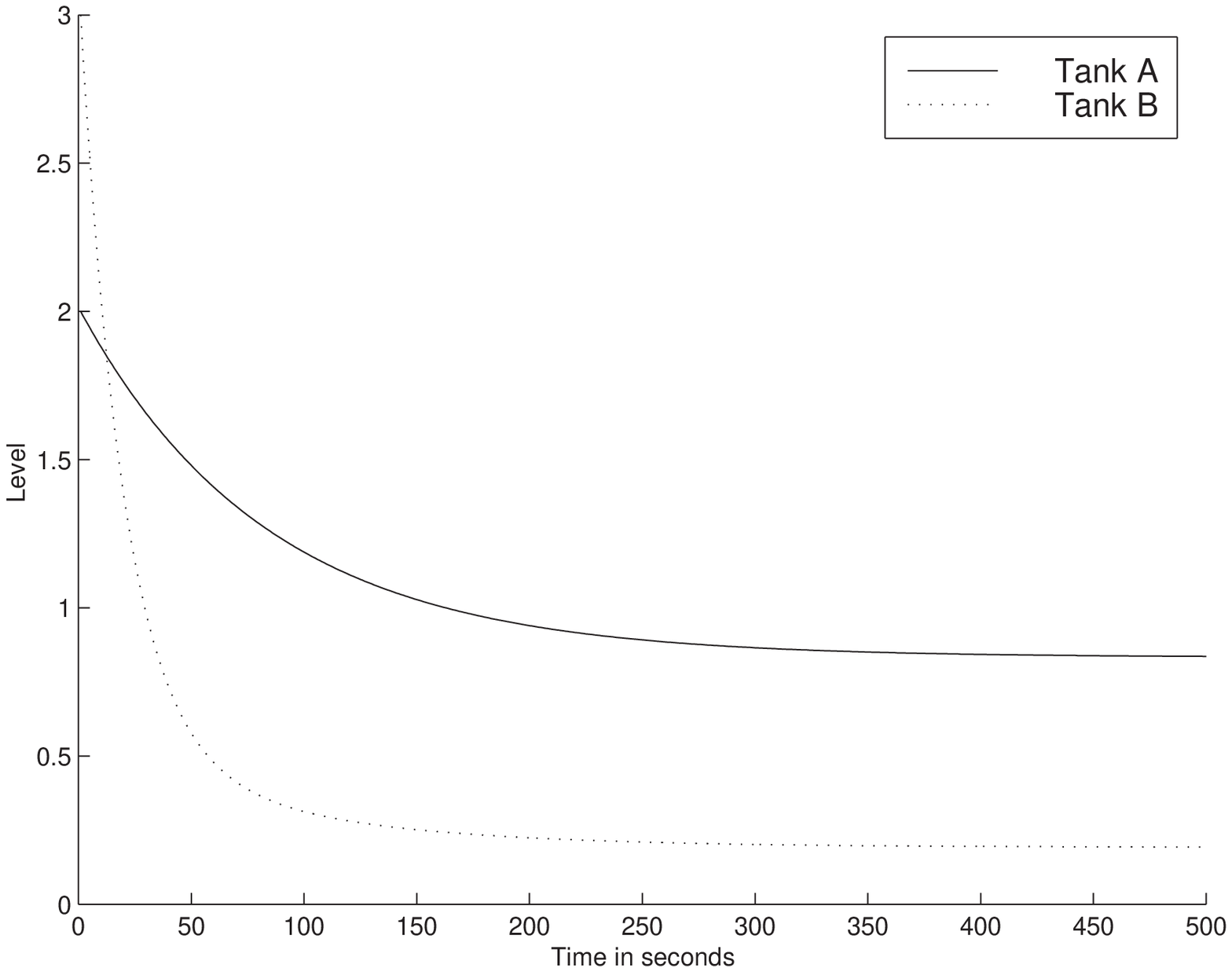}{1.0}
\end{minipage} \hfill
\begin{minipage}{65mm}
	\postscript{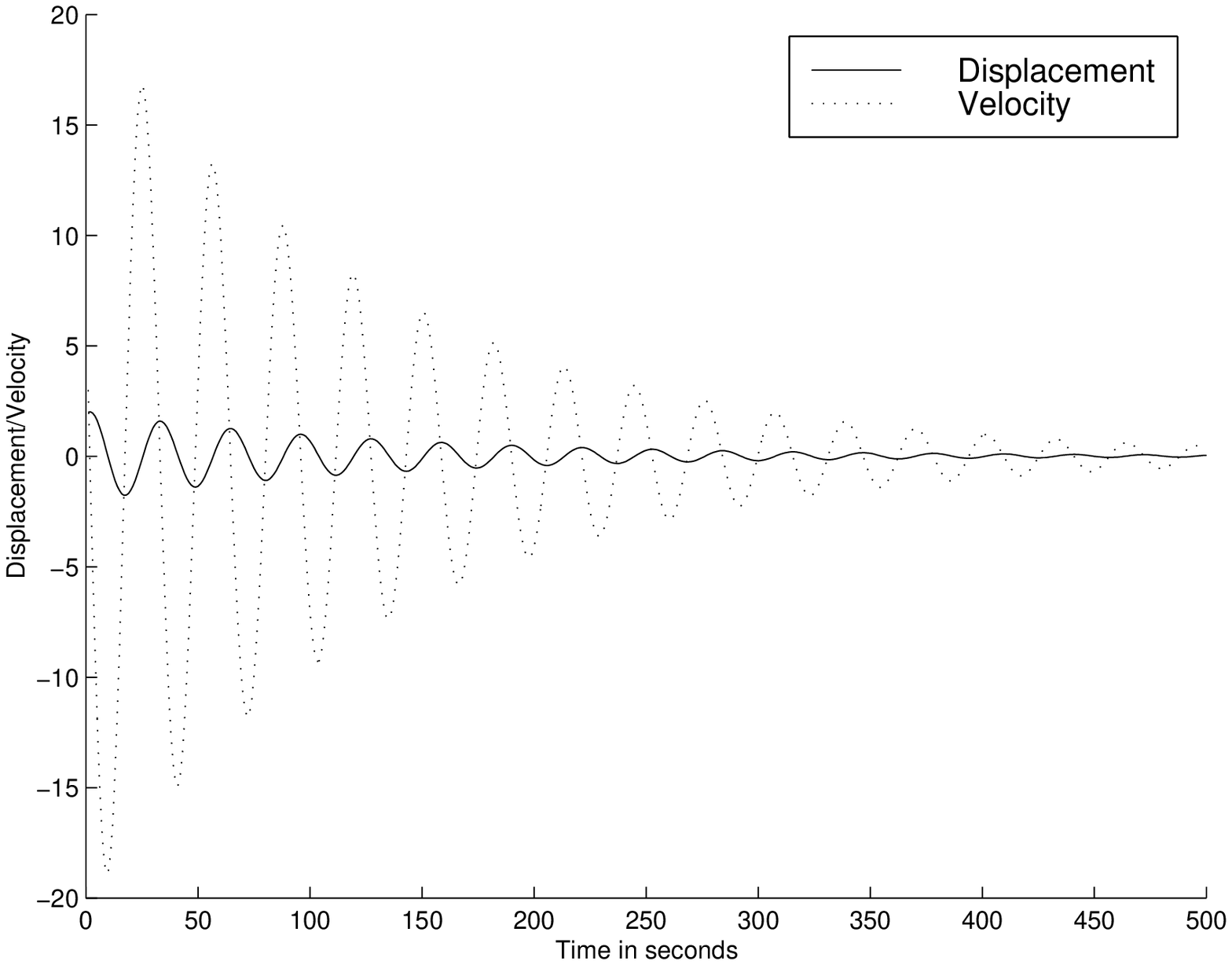}{1.0}
\end{minipage}
\caption{A graph of example numeric behavior of (a) the u-tube, top left; 
	(b) the coupled tanks, top right; (c) the cascaded tanks, bottom
	left, and (d) damped spring, bottom right.}
\label{nums}
\end{center}
\end{figure}

The models used for the numerical simulations had the same structure as the qualitative models, but with the substitution of a real valued parameter for each monotonic function relation. This gives a
linear relation between the two variables; more complex, non-linear
functions might have been used, but linear functions provided a suitable
approximation of the known behavior of these systems, as shown graphically in
Fig.~\ref{nums} (a)--(d); which is as much as is required for this  proof-of-concept study. 

For a given set of function parameter values, initial conditions, and input
value, a quantitative model produces a single quantitative behaviour
(this contrasts with qualitative models that can produce a list of all
possible behaviours for the model).  Parameter values were chosen so that the
models approached a steady state during the time period of the test.  
The models were implemented in Matlab 5.3 using the ODE15s ODE solver.  
%It was assumed that 
Each time point generated by the simulation was made available as part of the sampled data. This ensures that the sampling rate is suitably fast with respect to the Nyquist criterion. It also guarantees that a sufficient number of data points are available as required by the Beckman filter.

\subsubsection{Method}

{\bf Noise-free data.} We use the following method for
evaluating {\sc ILP-QSI}'s system-identification performance
from noise-free data:

\begin{enumerate}
\item[] For each of the four test-systems:
\begin{enumerate}
\item Obtain the system behaviour of the test system with
	a number of different initial conditions and input
	values. Convert each of these into qualitative states
	using the procedure in Section \ref{appb}.
\item Using {\em all\/} the qualitative states obtained as
        training data construct a set of models using
        {\sc ILP-QSI} and record the precision of the result
        (this is the proportion
        of the models in the result that are equivalent to the correct
        model). Thus, for each training data set,
        the result returned by {\sc ILP-QSI} will have
        a precision between $0.0$ and $1.0$.
\end{enumerate}
\end{enumerate}

\noindent
The following details are relevant:
(a) The quantitative models were
each put into three separate initial conditions, so that the magnitude of the two state variables were
set to $(2,0)$, $(0,3)$ and $(2,3)$.  Specifically, these were
the initial values of the two tank levels for all three tank systems,
and the displacement and velocity for the spring.
These values were not crucial but were
chosen for the initial conditions because they caused the numerical models to
converge on the steady state for each system in a reasonable number of
iterations;
(b) Each initial condition gave rise to a system behaviour and hence
a set of qualitative states. In the second step above, qualitative
states from {\em all} behaviours is used as training data. This is because kernel subsets 
necessary for correct system identification usually contain
qualitative states from multiple quantitative behaviours.
(c) The conversion process results in erroneous qualitative states
(see Section \ref{appb}). Thus, the training data used here 
contain both false positives and false negatives.

\vspace*{0.5cm}
\noindent
{\bf Noisy data.} We use the following method for
evaluating {\sc ILP-QSI}'s system-identification performance
from noisy quantitative data:

\begin{enumerate}
\item[] For each of the four test-systems:
\begin{enumerate}
\item Obtain the system behaviour of the test system with
        a number of different initial conditions and input
        values.
\item Corrupt each system behaviour with additive noise;
\item Convert each corrupted behaviour into qualitative states
        using the procedure in Section \ref{appb}.
\item Using {\em all\/} the qualitative states obtained as
        training data construct a set of models using
        {\sc ILP-QSI} and record the precision of the result
        (this is the proportion
        of the models in the result that are equivalent to the correct
        model). Thus, for each training data set,
        the result returned by {\sc ILP-QSI} will have
        a precision between $0.0$ and $1.0$.
\end{enumerate}
\end{enumerate}

\noindent
In the second step
noise was added to the numerical data sets as follows.
A Gaussian noise signal (with a $\mu$ of 0.0 and $\sigma$ of 1.0) was generated using by the built-in Matlab
function {\tt normrnd} and
scaled to three orders of magnitude of the original noise, namely 0.01, 0.1
and 1.0 (representing ``low'', ``medium'' and ``high'' amounts of
noise respectively). These scaled noise variants were added to the numerical
values of the system behaviour obtained from each initial condition.
	
\subsection{Quantitative Experimental Results}

	The process of converting from quantitative to qualitative states introduces errors, even for noise free data. Table~\ref{noiseTable} shows the proportion of correct qualitative states to the total number of qualitative states that were obtained from the numerical
signal, including noisy states.  The table shows this proportion for all four
systems, under the different degrees of noise.  The numerical simulations were not intended to be exhaustive and do not cover every possible such behaviour; so it is not surprising to observe that there is no case where all the states of the complete envisionment are generated. 

	The results of the qualitative experiments detailed in the previous section indicate that in order successfully to learn the target model data from all branches of the envisionment are required, and the greater the number of such states used the greater the liklihood of learning the target model structure. Therefore in these experiments we utilised all the states generated from the numerical simulations.

\begin{table} 
 \begin{small}
	\begin{center} \vspace{1ex}
	\begin{tabular}{|r|l||c|c|c|} \hline \hline
  \multicolumn{2}{|c||}{ } &
  	\multicolumn{3}{|c|}{Initial States} \\ \hline
             Model & Noise level &(2,0)&(0,3)&(2,3)\\ \hline \hline
	    	u-tube   & 0    & 4/6 & 4/6 & 3/5 \\ \cline{2-5}
	  	         & 0.01 & 2/8 & 2/8 & 2/9 \\ \cline{2-5}
	  	         & 0.1  & 2/10& 2/10& 2/15\\ \cline{2-5}
	  	         & 1    & 2/37& 2/37& 2/53\\ \hline
	  	Coupled  & 0    & 6/14& 5/13& 5/14\\ \cline{2-5}
		         & 0.01 & 6/16& 4/14& 4/12\\ \cline{2-5}
		         & 0.1  & 6/16& 5/25& 4/15\\ \cline{2-5}
		         & 1    & 6/58& 6/61& 4/46\\ \hline
		Cascaded & 0    & 5/8 & 5/8 & 3/8 \\ \cline{2-5}
		         & 0.01 & 4/10& 4/12& 2/9 \\ \cline{2-5}
		         & 0.1  & 4/17& 4/21& 2/13\\ \cline{2-5}
		         & 1    & 4/39& 4/49& 4/38\\ \hline
		Spring   & 0   & 20/33&19/35& 22/39\\ \cline{2-5}
		         & 0.01& 23/48&20/36& 18/41\\ \cline{2-5}
		         & 0.1 & 23/49&20/38& 18/44\\ \cline{2-5}
		         & 1   & 20/65&20/52& 19/53\\ \hline \hline
	\end{tabular}
	\end{center} \end{small}
		\caption{The input data for the numeric experiments, described
		as the proportion of the number of clean states / the total
		number of converted states for different systems and degrees
		of noise.}
	\label{noiseTable}

\end{table}

The results from the numerical data experiments are shown in
Fig.~\ref{numExps}.  These experiments show that it is possible to learn models from clean
and noisy numerical data even when the qualitative states generated from the
clean numerical data contain a number of unavoidable data transformation
errors.  The results for each of the systems used are as follows:

\begin{figure}[htb]
  \includegraphics[scale=0.7]{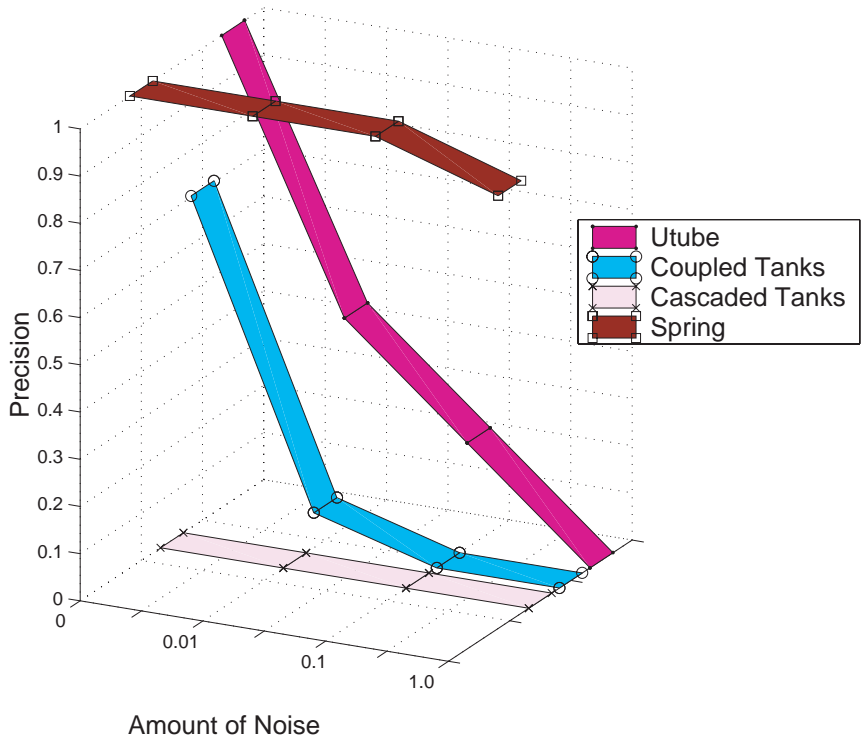}
  \caption{A comparison of the results from the numerical learning
  	tests, averaged over all three initial conditions.
  	Tests that attempted to learn from few states, such as the
	cascaded tanks, are more likely to fail than those that have
	large numbers of states, such as the spring.  This is consistent
	with the kernel subset principle introduced in
	Section~\ref{exp} since model-learning is not
	precise without the presence of certain key states in the input
	data.}
  \label{numExps}
\end{figure}
	
\begin{description}

\item[The spring system]  This system has 31 states in the complete envisionment and Table \ref{noiseTable} shows that the quantitative to qualitative conversion process yields around 20 of those states. It can be seen from Fig. \ref{gr4} that learning from 20 states gives 100\% precision in learning this target model, even in the presence of noise. It is not surprising, therefore, to find that the learning precision is perfect up to the highest noise level. Since the qualitative experiments were done by sampling, the slight  downturn at the  highest noise level could be due to the large number of noisy states generated in this experiment. Hence we can say that these results are in keeping with the qualitative experiments.

\item[u-tube \& coupled tanks] The complete envisonments for these systems contain 6 and 10 states respectively. Table \ref{noiseTable} shows that the number of true states generated is less than the complete envisionment, and significantly less than the number of noisy states in each case. As one would expect from the results presented in Fig. \ref{ut2d} the u-tube gives better results than the coupled tanks (having a higher proportion of the envisionment states present). Ultimately, the ability to learn the model is completely curtailed by the noise; though sooner in the case of the coupled tanks (which the qualitative experiments show to be more sensitive to the presence of noise). This is consistent with the results of the qualitative experiments.

\item[The cascaded tanks] In the qualitative learning experiments all the kernel subset triples had state 0 included. This is the state representing the situation where both tanks are empty to begin with, and is not one of the initial states included in the numerical simulations. Also, a perusal of Fig. \ref{cs1} reveals that the introduction of noise radically reduces the learning precision, and for 4 states (the average number of true states generated by the qualitative to quantitative conversion process) it is zero. Taking account of all these facts it was to be expected that the cascaded tanks model would not be successfully identified by these experiments. This is again consistent with the findings of the qualitative experiments.

\end{description}

\section{Application to Biological System Identification}
\label{glycExps}

The work reported thus far has been aimed at demonstrating the viability of ILP-QSI and of identifying the bounds of operation of the approach. In this section we examine scalability of the {\em method} to identify a complex real world biological network. We use the glycolysis pathway as a test case for identification.

\subsection{The Test System: Glycolysis}

We chose to study the metabolic pathway of glycolysis as a test case.   
Glycolysis is one of the most important and ubiquitous in biology, it was 
also historically one of the first to be discovered, and 
still presents a challenge to model accurately.

The QSIM  primitives are sufficient
to model adequately the qualitative behaviour of the glycolysis pathway.
There are, however, two problems. First, few biologists would understand
such a model, as he or she would reason at a much higher level
of abstraction.
Second, the computational complexity of the corresponding system
identification task for glycolsis (a qualitative model with 100 or more
QSIM relations) is, at least currently, intractable.
We address both these by modelling metabolic
pathways in a more abstract manner using biologically meaningful
\textit{metabolic components} (MC) \cite<a similar approach to
constructing complex qualitative models of the human heart was used
in>{kardio}.  Specifically, we note
that in metabolic pathways, there are essentially two
types of object: metabolites (small molecules) and
enzymes (larger molecules that catalyze reactions). We use component
models of each of these objects as described below \cite{king05}.

\subsubsection{Modelling Metabolites and Enzymes}

The concentrations of metabolites vary over time as they are
synthesised or utilised by enzymatically catalysed reactions. As a
result their concentration at any given time-point is a function of:
(a)  their concentration at the previous time-point; and
(b) the degree to which they are used or created by various enzyme reactions.

When modelling enzymes, each enzyme is assumed to have at most two
substrates and at most two products. If there are two substrates or products
these are considered to form a substrate or product complex, such that
the amount of the complex is proportional to the amount of the
substrates or products multiplied together. This models
the probability that both substrates (or products) will collide with
the enzyme with sufficient timeliness to be catalysed into the product
complex (or substrate complex). The substrate complex is converted
into the product complex, which then disassociates into the product
metabolites, and vice versa. We shall use the 
phrase ``flow through the enzyme'' to denote the amount of substrate complex formed minus the amount of product
complex formed. (See the work of \citeA{voit00} for details of enzyme kinetics.)

\begin{figure}[hbtp]
\begin{center}
\includegraphics*[scale=0.75]{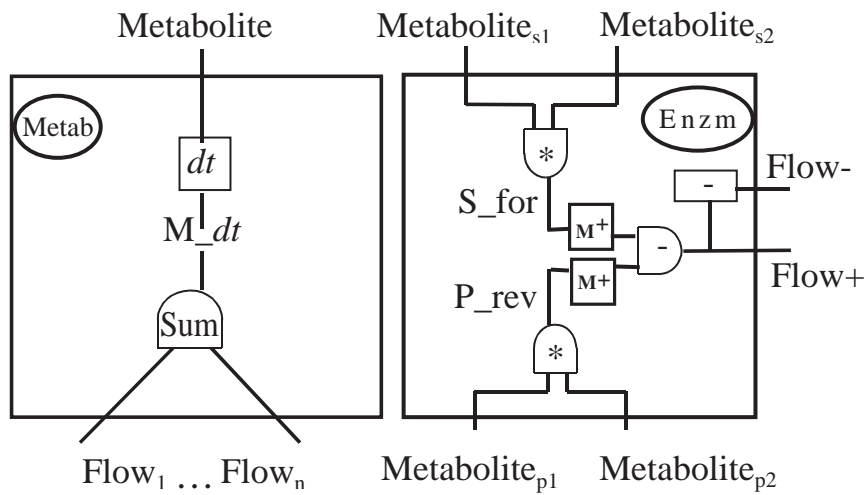}
\end{center}
\caption{\label{mc} The Metabolic Components (MCs)  used in the biological experiments, with the underlying QSIM primitives.}
\end{figure} 

Quantitative, and corresponding qualitative representations of the
metabolite and enzymes using QSIM relations, are therefore:
\\[2mm]
\begin{small}
\begin{minipage}{45mm}
  Metabolites:
  \begin{equation}\label{metab}
     \frac{dM}{dt} = \sum_{i=1}^n Flow_i
  \end{equation}
  {\tt DERIV($Metabolite$,$M_dt$),\\
       SUM($Flow_1$, $\dots$, $Flow_n$, $M_dt$).\\\\
	}
\end{minipage} \hfill
\begin{minipage}{100mm}
  Enzymes:
  \begin{equation}
     Flow_i =f( \prod_{s=1}^{S} Metabolite_s) - g( \prod_{p=1}^{P} Metabolite_p)
  \end{equation}
  {\tt PROD($Metabolite_1$, $\ldots$, $Metabolite_s$, {\it S-for}),\\
       PROD($Metabolite_1$, $\ldots$, $Metabolite_p$, {\it P-rev}),\\
       M+({\it S-for}, Ds), \\
       M+({\it P-rev}, Dp),\\
       SUB(Ds, Dp, $Flow$), \\
       MINUS($Flow$,$Flow_{minus}$).}
\end{minipage}
\end{small}\\ \\

\noindent
Here, $S$ refers to the input metabolites to an enzymatic reaction, or its
{\it substrates}, and $P$ refers to the products of an enzymatic reaction.
The {\tt SUM()} and {\tt PROD()} predicates are simply extensions of the
{\tt ADD()} and {\tt MULT()} predicates, over any number of inputs. Fig. \ref{mc} shows how these constraints are grouped together as metabolic components (MCs). This permits us to create more general constraints representing the {\it metabolite} and {\it enzyme} components as follows:

ENZYME(($S_1, S_2$) ($P_1, P_2$) $enzymeFlow$)

METABOLITE($metaboliteConc$ $metaboliteFlow$ ($enzymeFlow_1 \ldots enzymeFlow_n$))

Here the {\tt ENZYME} predicate
identifies the substrates and products (the first argument) and
returns a single variable representing the
flow through the enzyme (the second argument). The {\tt METABOLITE} predicate relates
the level and flow of metabolites (the first and second
arguments) with the flow through enzymes (the third argument).

\subsubsection{Modelling Glycolysis}

Using qualitative components representing metabolites and enzymes, 
we construct a qualitative model of glycolysis. 
Our model uses 15 metabolites, namely:
pyruvate (Pyr), glucose (Glc), phosphoenolpyruvate (PEP),
fructose 6-phosphate (F6P), glucose 6-phosphate (G6P),
dihydroxyacetone phosphate (DHAP), 3-phosphoglycerate (3PG),
1,3-bisphosphoglycerate (13BP), fructose 1,6-biphosphate (F16BP),
2-phosphoglycerate (2PG), glyceraldehyde 3-phosphate (G3P), ADP, ATP,
NAD, and NADH. We have not included H+, H$\mbox{}_\mathrm{2}$O, or
Orthophosphate as they are assumed to be ubiquitous (in addition,
the restriction of substrates and products to being at most three in number
prevents their inclusion).

\begin{figure}
{\small{
\begin{center}
\begin{tabular}{r l l} 
1. & ({\it Hexokinase}): & Glc + ATP $\rightarrow$ G6P + ADP. \\
2. & ({\it Phosphoglucose isomerase}): & G6P $\rightarrow$ F6P. \\
3. & ({\it Phosphofructokinase}): & F6P + ATP $\rightarrow$ F16BP + ADP \\
4. & ({\it Aldolase}): & F16BP $\rightarrow$ DHAP + G3P \\
5. & ({\it Triose phosphate isomerase}): & DHAP $\rightarrow$ G3P \\
6. & ({\it Glyceraldehyde 3-phosphate dehydrogenase}): & G3P + NAD $\rightarrow$ 13BP + NADH. \\
7. & ({\it Phosphoglycerate kinase}): & 13BP + ADP $\rightarrow$ 3PG + ATP. \\
8. & ({\it Phosphoglycerate mutase}): & 3PG $\rightarrow$ 2PG. \\
9. & ({\it Enolase}): & 2PG $\rightarrow$ PEP \\
10. & ({\it Pyruvate kinase}): & PEP + ADP $\rightarrow$ Pyr + ATP.\\ 
\end{tabular} 
\end{center}
}}
\caption{The reactions included in our qualitative model of glycolysis.
The reactions that consume ATP and NADH are not explicitly included.}
\label{fig:glycTab}
\end{figure}
The qualitative state of glycolysis is defined by the set of
qualitative states of the 15 metabolites.  Table~\ref{glycstate}
is a representation of one
such qualitative state.  To understand this state consider the
first entry intended to represent the qualitative state of NAD
(that is, NAD concentration: $<0, \infty), dec>$, and NAD flow: $<(-\infty, 0), dec>$).
The meaning of this is that the concentration of NAD  is positive $(0, \infty)$
and decreasing (dec), and the rate of change of the concentration of NAD (in
analogy to the physical systems, the ``flow'' of NAD)
is negative $(-\infty, 0)$ and decreasing (dec).  Similar meanings apply
to the other metabolites.  Note that metabolic concentrations
must be between 0 and $\infty$; it cannot be negative, and the
0 state is uninteresting.
\begin{center}
\begin{table*}
\begin{tabular}{|l|l|l|}
\hline
Metabolite & Concentration & Flow \\
\hline
NAD & $<(0, \infty), dec>$  & $<(-\infty, 0), dec>$ \\
NADH  & $<(0, \infty), inc>$ & $<(0, \infty), inc>$ \\
ATP &  $<(0, \infty), dec>$  & $<(-\infty, 0), dec>$ \\
ADP & $<(0, \infty), dec>$  & $<(-\infty, 0), dec>$ \\
Pyr & $<(0, \infty), inc>$ & $<(0, \infty), dec>$ \\
Glc & $<(0, \infty), dec>$  & $<(-\infty, 0), inc>$ \\
PEP & $<(0, \infty), dec>$  & $<(-\infty, 0), dec>$ \\
F6P & $<(0, \infty), dec>$  & $<(-\infty, 0), dec>$ \\
G6P & $<(0, \infty), dec>$  & $<(-\infty, 0), dec>$ \\
DHAP & $<(0, \infty), dec>$  & $<(-\infty, 0), dec>$ \\
3PG & $<(0, \infty), inc>$ & $<(0, \infty), std>$ \\
13BP & $<(0, \infty), std>$ & $<0, inc>$ \\
F16PB &  $<(0, \infty), inc>$ & $<(0, \infty), dec>$ \\
2PG & $<(0, \infty), dec>$  & $<(-\infty, 0), dec>$ \\
G3P & $<(0, \infty), inc>$ & $<(0, \infty), inc>$ \\
\hline
\end{tabular}
\caption{\label{glycstate}
A  legal qualitative state of the 15 metabolites observed during glycolysis.}
\end{table*}
\end{center}
Using this representation, a possible model for
glycolysis is shown in Fig.~\ref{fig:glycmodel}.
The model describes constraints on the levels and
``flows'' of metabolites. 
Thus, the constraint
{\tt enzyme((G3Pc, NADc), (13BPc, NADHc), Enz6f)} states that the flow
through enzyme 6 (Enz6f) controls the transformation of the concentrations
G3Pc and NADc into the levels 13BPc and NADHc; whereas the constraint
{\tt metabolite(NADc, NADc, (Enz6f, -))} states that the concentration (NADc) and
flow (NADf) of the metabolite NAD is controlled by flow through the
single enzyme number 6 (Enz6f : Glyceraldehyde 3-phosphate
dehydrogenase), and that this enzyme removes (signified by the `-') NAD
(`+' would mean the enzyme flow adds the corresponding metabolite).
\begin{figure*}
{\scriptsize{
\begin{verbatim}
      ENZYME((Glcc, ATPc),(G6Pc,ADPc),Enz1f),
      ENZYME((G6Pc),(F6Pc),Enz2f),
      ENZYME((F6Pc,ATPc),(F16BPc,ADPc),Enz3f),
      ENZYME((F16BPc),(G3Pc,DHAPc),Enz4f),
      ENZYME((DHAPc),(G3Pc),Enz5f),
      ENZYME((G3Pc,NADc),(13BPc,NADHc),Enz6f),
      ENZYME((13BPc,ADPc),(3PGc,ATPc),Enz7f),
      ENZYME((3PGc),(2PGc),Enz8f),
      ENZYME((2PGc),(PEPc),Enz9f),
      ENZYME((PEPc,ADPc),(Pyrc,ATPc),Enz10f),
      METABOLITE(ATPc,ATPf, (Enz10f +), (Enz7f, +),(Enz1f, -),(Enz3f, -)),
      METABOLITE(ADPc,ADPf,(Enz1f, +),(Enz3f, +),(Enz10f, -)(Enz7f, -)),
      METABOLITE(NADc,NADf,(Enz6f, -)),
      METABOLITE(NADHc,NADHf,(Enz6f, +)),
      METABOLITE(Pyrc,Pyrf,(Enz10f, +)),
      METABOLITE(Glcc,Glcf,(Enz1f, -)),
      METABOLITE(PEPc,PEPf,(Enz9f, +),(Enz10f, -)),
      METABOLITE(F6Pc,F6Pf,(Enz2f, +),(Enz3f, -)),
      METABOLITE(G6Pc,G6Pf,(Enz1f, +),(Enz2f, -)),
      METABOLITE(DHAPc,DHAPf,(Enz4f, +),(Enz5f, -)),
      METABOLITE(3PGc,3PGf,(Enz7f, +),(Enz8f, -)),
      METABOLITE(13BPc,13BPf,(Enz6f, +),(Enz7f, -)),
      METABOLITE(F16BPc,F16BPf,(Enz3f, +), (Enz4f, -)),
      METABOLITE(2PGc,2PGf,(Enz8f, +),(Enz9f, -)),
      METABOLITE(G3Pc,G3Pf,(Enz5f, +),(Enz4f, +),(Enz6f, -)).
\end{verbatim}
}}
\caption{A  representation of a qualitative model of glycolysis
(see text for details).}
\label{fig:glycmodel}
\end{figure*}

\subsection{Experimental Aim}

The specific system identification task we were interested in is:
Given qualitative observations of metabolic states,
can {\sc ILP-QSI} identify a correct qualitative model for glycolysis?

\subsection{Materials and Method}
\begin{figure}[h]
\begin{center}
\includegraphics*[scale = 0.7]{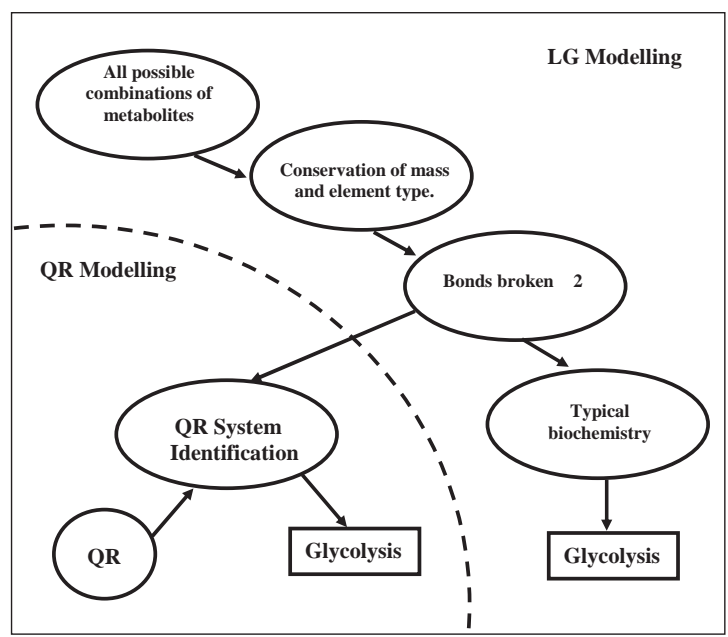}
\end{center}
\caption{\label{glycmethod} Our Metabolic System Identification methodology.}
\end{figure}

Our methodology is
depicted in Fig.~\ref{glycmethod}, where we describe two separate ways of
identifying biochemical pathways.  We make the following assumptions:
\begin{enumerate}
\item
The data are sparse and not necessarily measured as part of a continuous time series. This is realistic given current experimental limitations in metabolomics, and rules out the possibility of numerical system identification approaches. 
\item
Only metabolites of known structure are involved in the model. The reason for this is that we employ a chemoinformatic heuristic to 
decrease the number of possible reactions. The heuristic is based on 
the reasonable assumption that any chemical reaction catalysed by an 
enzyme only breaks a few chemical bonds. Full details are in the paper by \citeA{king05}. This is the strongest assumption we make. Even given the rapid advance of metabolomics (NMR, mass-spectroscopy, etc.), it is not currently realistic to assume that all the relevant metabolites in a pathway are observed and their structure determined. 
\item
Only metabolites of known structure are involved in a particular pathway. This is a restriction because current metabolomics technology can observe more compounds than can be structurally identified.
\item
All reactions involve at most three substrates and three products.
\item
For the qualitative states: we can measure the direction of change in 
metabolite level and first-derivative. This requires sampling the level 
at least three times in succession.

\end{enumerate}

\subsubsection{Logical/Graph-based Constraints}

We first considered the logical/graph-based (LG) nature of the problem. The
specific domain of metabolism imposes strong constraints on possible
LG based models. We used these constraints in the following way:
\begin{enumerate}
\item
Chemical reactions conserve matter and atom type \cite{valdesPerez94}. For glycolysis we generated all possible ways of combining the 18 metabolites to form matter and atom type balance reactions ($\leq$ 3 reactants and $\leq$ 3 products). This produced 172 possible reactions where the substrates balanced the products in the number and type of each element. The number compares well with the ~2,300,000 possible reactions which would naively be possible.
\item
Typical biochemical reactions only make/break a few bonds, and cannot arbitrarily rearrange atoms to make new compounds. A reaction was considered plausible if it broke 1 bond per reactant.  This analysis was done originally by hand, and we have subsequently developed a general computer program that can automate this task. Of the 172 balanced reactions 18 were considered chemically plausible. Of these 18  reactions, 10 are the actual reactions of glycolysis and  8 are decoy reactions.
\end{enumerate}

\subsubsection{Qualitative Reasoning Constraints}

We used a simple generate and test approach to learning. For the
first computational experiment we used the 10 reactions of glycolysis
and the 8 decoy reactions that were considered chemically feasible,
see Fig.~\ref{fig:glycTab}. All these reactions, in the absence of evidence to the
contrary, are considered to be irreversible.  We first generated all
possible ways of combining the 18 reactions which connected all the 15
main substrates in glycolysis (models are non-disjoint). This
generated 27,254 possible models with $\leq$ 10 reactions - it was not
necessary to look for models with more reactions than that of the
target (parsimony), as the models can be generated in size order. The
smallest number of reactions necessary to include all 15 metabolites
was of size 5. All of the 27,254 models involved the reaction:
glyceraldehyde 3-phosphate + NAD $\Leftrightarrow$ 1,3-bisphosphoglycerate + NADH
(reaction 6); so we could immediately conclude that this reaction
occurred in glycolysis.

We formed example qualitative states of glycolysis using our QR
simulator (in a pseudo random manner) to test these models. The states thus generated did not contain any noise. The 27,254
possible models were then tested against these states, and if a model
could not generate a particular state it was removed from
consideration (accuracy constraint).  Note that the
flows of the metabolites through each enzyme are not observed - they
are intermediate variables.  All we observe are the overall levels and
flows of the metabolites.  This makes the system identification task
much harder.

For efficiency, we used the fast YAP Prolog compiler.  We also
formed compiled down versions of the enzyme and metabolite MCs
(input/output look-up tables), and compiled down parts of QSIM.  We
also adopted a resource allocation method that employed increasingly
computationally expensive tests: i.e. forming filter tests with
exponentially increasing numbers of example states.

\subsubsection{Results}

After several months of compute time on a 65 node Beowulf cluster we
reduced the 27,254 possible models down to 35 (a ~736 fold
reduction). These models included the target model (glycolysis), plus
34 other models that could not be qualitatively distinguished from it.
All 35 models included the following six reactions (see  Fig. \ref{fig:glycTab}): 

\noindent
3.	F6P + ATP $\rightarrow$  F16BP + ADP \\ 
4.	F16BP $\rightarrow$ DHAP + G3P \\	 
5.	DHAP $\rightarrow$ G3P \\
6.	G3P + NAD $\rightarrow$ 13BP + NADH \\
8.	3PG $\rightarrow$ 2PG \\
9.	2PG $\rightarrow$ PEP \\
\emph{These reactions form the core of glycolysis.}

Examining the 35 models also revealed that the correct model had the fewest cycles, however we do not know if this is a general phenomenon.  

We attempted to use the Progol positive only compression measure to
distinguish between these models.  This is based on comparing the
models on randomly generated states.  However, this was unsuccessful
as no model covered any of the 100,000 random states we generated!  We
believe this is due to the extremely large state space.  However, a
simple modification of this approach does work.  If we produce random
states from glucogenesis (glycolysis driven in the reverse direction),
then the true model of glycolysis covers fewer examples than any of the 34
alternatives, and so is identified as the true model. Note that this approach, unlike the use of Progol positive only compression measure, requires that new experimental data is obtained.

Thus we have demonstrated that the {\em method} for learning qualitative models of dynamic systems is scalable to handle a relatively large metabolic system. We have achieved this by means of MCs that represent meaningful units in the domain, and which map directly to the QSIM constraints from which they are abstracted. They also enable us to present these more complex models in a more ``user friendly'' manner, removing the need to understand the structure of high order differential equations.

\section{Related Work}
\label{relwork}
	
System identification has a long history within
machine learning: we present below some of the important signposts
that were directly relevent to the work here. {\it We focus on the strand of research which deals with learning qualitative models of dynamic systems.}

The earliest description of which we are aware concerning a computer program 
identifying a quantitative model to explain experimental data is the
work by \citeA{coll:mi2}. There a procedure is described
that heuristically searches through equation structures,
which are linear combination of functions of the observed variables.
Better known is the
{\sc Bacon} system \cite{LangleyP81},
early versions of which
largely concentrated on the parameter estimation problem, in particular selecting the most appropriate values for any exponents in the equations.
For example, given a class of algebraic equation structures
{\sc Bacon}.1 was able to reconstruct Kepler's model for planetary
motion from data.  While later work \cite<for example the work of>{Nordhausen93}
attempted to extend this work to deal with identifying
both the algebraic structure and the relevant parameters,
{\sc Bacon} highlighted the importance
of {\it bias} \cite{MitchellT86} in machine learning, both
in constraining the possible model structures and
in the space of possible models conforming to those structures.
Other quantitative equation discovery systems in this lineage
are: {\sc Coper} \cite{kokar:coper}, that uses
dimensional analysis to restrict the space of equations;
{\sc Fahrenheit/EF} \cite{langley:fahrenheit} and {\sc E}$\mbox{}^{*}$
that only examine the space of bivariate equations; {\sc Abacus}
\cite{falkenhainer:abacus} that can identify piecewise
equations; {\sc Sds} that uses type and dimensionality restrictions
to constrain the space of equations; the {\sc Lagrange} family of equation finders
\cite{Dzeroski93,Todorovski97,Todorovski00,Todorovski03} which
attempt to identify models in the form of ordinary and
partial differential equations; and IPM \cite{langley03} with its extensions and developments, Prometheus/RPM \cite{bridewell04,asgharbeygi05}, which incorporate process descriptions \cite{Forbus84} to aid the construction and revision of quantitative dynamic models.

Focussing specifically on non-classical system identification
for metabolic models, perhaps the most notable work on identification
is that of \citeA{arkin:science} who identified a graphical model of the
reactions in a part of glycolysis from experimental data.
The work of \citeA{reiser:yeast}
presents a unified logical approach to simulation (deduction)
and system identification (induction and abduction).
An interesting recent approach, presented by \citeA{kosa:pathway},
examines the identification of metabolic ODE models using genetic programming techniques. In this, the cellular system is viewed as an electrical circuit and
the space of possible circuits is searched by means of a genetic programming
approach.

The earliest reported work on the identification of qualitative models
is that of \citeR[and colleagues,]{Mozetic87}
who identified a model of the electrical activity of the heart.
This work, reported more fully in \cite{kardio} remains a landmark
effort in the qualitative modelling of a complex biological system.
However, as other researchers have noted \cite{Muggleton91},
these results were obtained only for static models and
did not provide insight into how models of dynamic systems should be identified.

The first machine learning system for learning qualitative models of
dynamic systems was {\sc Genmodel}
\cite{Coiera89,Coiera89a}.  {\sc Genmodel} did
not need any negative examples of system behaviour and models learned were
restricted to qualitative relationships amongst the observed
variables only (that is, no intermediate, or hidden, variables
were hypothesized). The model, obtained using the
notion of a most specific generalization of observed variables
\cite<in the sense of>{plotkin:thesis}, was usually over-constrained.
That is, it contained more constraints than necessary
to characterize fully the dynamics of the
system being modeled.
An updated version of {\sc Genmodel} developed by \citeA{Hau93} showed that dimensional analysis
\cite{Bhaskar90} could be used as a form of directed negative example
generation.  The new version could learn from 
from real-valued experimental data (which were converted internally
into a qualitative form), but still required all variables to be known and
measured from the outset.
The system {\sc MISQ}, entirely similar in complexity
and abilities to the earlier version of {\sc Genmodel} was developed by
\citeA{Kraan91}. This was later
re-implemented in a general-purpose relational learning program Forte \cite{forte}, which allowed the hypothesis of intermediate
variables \cite{richards:qm}. The relational pathfinding approach used by {\sc MISQ} (through
the auspices of Forte) is a special form of Inductive Logic Programming, the general framework of which is much more powerful

Bratko and colleagues were the first to view the problem of learning
dynamic qualitative models explicitly as an exercise in
Inductive Logic Programming (ILP)
and first demonstrated the
possibility of introducing intermediate (unobserved)
variables in the models.  They used the ILP system {\sc Golem}
\cite{Muggleton90} along with the QSIM representation
to produce a model of the u-tube system.
The model identified was equivalent to the accepted model
(in the sense that it \textit{predicted} the same behaviour)
but the structure generated was not in a form that could help \textit{explain} the
behaviour \cite{Coghill01}. Like {\sc Genmodel}, the
model produced was over constrained.
Unlike {\sc Genmodel}, {\sc Golem} required both positive and
negative examples of system behaviour and was shown by \citeA{Hau93} to be sensitive
to the actual negative examples used. 

\citeA{SayKuru96} describe a program for 
system identification from qualitative data called {\sc QSI}.
{\sc QSI} first finds correlations between variables,
and then iteratively introduces new relations (and intermediate
variables), building a
model and comparing the output of that model with the known states until a
satisfactory model is found.  Say and Kuru characterized this approach as one
of ``diminishing oscillation'' as it approaches the correct model.
Like {\sc Genmodel} and {\sc MISQ}, {\sc QSI} does not require
``negative'' observations of system behaviour. Unlike those systems,
it does not use dimensional analysis and there does not appear to
be any mechanism of incorportating such constraints easily within the
program. The importance of dimensional analysis is recognised though: the
authors suggest that it should be central to the search procedure. 

Thus, while the identification of quantitative models has had
a longer history in machine learning, learning qualitative models has
also been the subject of notable research efforts. In our view,
{\sc MISQ} (the version as implemented within Forte) and
{\sc QSI} probably represent
the state-of-the-art in this area. Their primary shortcomings 
are these:

\begin{itemize}
\item[--] It is not apparent from the description or experimental
	evaluation of {\sc MISQ} whether or not it is able to
	handle imperfect data (the correctness theorem presented
	only applies for complete, noise-free data).
\item[--] {\sc MISQ} seeks the most constrained model that is consistent
	with the data. Often, exactly the opposite is sought (that is,
	we want the most parsiomonius model).
\item[--] {\sc QSI} only deals with qualitative data and does not appear
	to include any easy mechanism for the incorporation of
	new constraints to guide its search.
\end{itemize}

\section{General Discussion}
\label{discussion}

In this paper we have presented a method for learning  qualitative models of dynamic systems from time-series data (both qualitative and quantitative). In this section we discuss the general findings and limitations, as well as suggesting a number of directions for developing this research theme.

\subsection{Computational Limitations}

A major limitation of the ILP-QSI system in identifying glycolysis was
the time taken (several months on a Beowulf cluster) to reduce the
models from the ~27,000 possible ones generated using chemoinformatic
constraints, to the single correct one using the qualitative state
constraints.  While it would be preferable for this process to be
faster, it is important to note that identifying a system with 10
reactions and 15 metabolites from scratch is an extremely hard
identification task.  We doubt that any human could achieve it, and we
believe it would be a challenge for all the system identification
methods we are aware of.  It is difficult to compare system
identification methods and we believe there is a need for competitions
such as those run
by KDD to compare methods. 

The computational time of identification is dominated by the time
taken to test if a particular model can produce certain observed
states: examining ~27,000 models is not unusual for a machine learning
program, but it is unusual for a program to take hours to test if
individual examples are covered.  The slow speed of our identification
method is therefore not a problem with what is normally considered the learning method (i.e. how the search of the space of possible models is done), but rather, is intrinsic to the
complex relationship between a model and the states it defines.  The 
cover-test method is, in the worst case, exponential in the maximum
size of the model.  Note that our
lack of an efficient, i.e. polynomial algorithm, to determine cover is
not because we are using qualitative states. We believe that the
inherent difficulty of this task applies to both quantitative and
qualitative models.  In some areas of mathematics moving from the
discrete to the real domain can simplify problems - this is the basis
of much of the power of analysis.  However, there is currently little
evidence that this is the case in system identification, and quantitative
models would seem to aggravate the problem.  As cover tests are
essentially deductions: can a set of axioms and rules (computer
program/model) produce a particular logical sentence (observed state);
they are in general non-computable.  However, in real scientific
systems, as they are bounded in space and time, non-computability is
not a problem, however we expect all system identification methods to
struggle with the task \cite{sadot.et.al.08}.  

\subsection{Kernel Subsets}

From our presentation of the results of the experimentation it is clear that certain subsets of states (termed the \textit{kernel subsets}) guarantee that the target model will be learned. From the analysis of these kernel subsets of state sets, we
hypothesise that the kernel sets reflect the qualitative structure of the system of interest.

For a coupled system, in order to learn the structure of a system with a high degree of precision, the data used should
come from tests yielding qualitatively different behaviors:
i.e.~behaviors which would appear as distinct branches in an envisionment
graph.  However, this hypothesis only provides a necessary, not a sufficient,
condition for learning because it does not identify which states in each branch are suitable starting points for an experiment.   For example consider the coupled tanks system. One could select states 9 and 11; these are from different branches yet they do not form a kernel subset. On the other hand, it was noted in presenting the results for this system that the key
states in these kernel subsets were states 7 and 8.  These states are in different branches and represent the {\it critical points} of the first derivative of the state variables of the system. This indicates the importance of these states to the definition of a system.

If a test were set up in which all the state variables were at their
critical points then the test could be run for a very short time and
the correct model structure identified.  However, it is probably impossible
in practice to set up such a test; especially in the situation where
the structure of the system is completely unknown.  An alternative is to set
up multiple tests with the state variables set to their extrema: from
which initial conditions all the states of the envisionment will eventually
be passed through. However it still may be difficult to set up such
tests, and they could take a long time to complete. These two
scenarios form the ends of a spectrum within which the most practical experimental
setting will lie.  The identification of the best strategies is an important
area of research to which the present work is clearly relevant.

On the other hand, for cascaded systems the kernel sets capture the asymmetry in the structure.  Here again the extrema and critical points play an important role; but in this case it is subordinate to the  fact that ILP-QSI automatically decomposes the system into its constituent parts for learning. This fact points to an important conclusion for learning larger scale complex systems; namely that the learning can be facilitated by, where possible, decomposing the system into cascaded subsystems.

\subsection{Future Work}
Having validated ILP-QSI on data derived from real biological systems,
the next step is to explore how successful it can be at modelling real
experimental data.  It would be relatively straightforward to obtain data from 
water tanks and springs, but it would be much more interesting to 
work on real biological data.  
For such work to be successful it is likely that the quantitative to
qualitative conversion process will need to be improved.  Although not
the focus of the work here, developing a more rigorous approach would
be crucial in using ILP-QSI in a laboratory setting \cite{biswas98}.  Once this
has been done it will be much easier to use real experimental data
for analysis by ILP-QSI.  Specifically, the improvements required
are the ability to extract all the qualitative states passed through
during a numerical simulation, whilst minimizing noise.  Nevertheless,
this is not a direct limitation of our ILP-QSI method.

The following possibilities would benefit from further investigation:
\begin{itemize}
	\item
The QR representation used could be changed from QSIM to a more
detailed and flexible one such as Morven \cite{Coghill94,Coghill96}.
	\item
The hypotheses presented about kernel subsets, such as why they are formed from
some states and not others, need to be confirmed and analyzed further.
	\item 
The ability to map and explore the features of the model space would be of
great use in planning further enhancements and, alongside kernel subsets,
will help give an understanding of exactly which states will allow reliable
learning. 

\item Large scale complex systems are generally identified piece by piece. The results from the cascaded tanks experiments indicate some circumstances under which this may be most easily facilitatied. Further investigation of this is warranted.

\item As an alternative to the methods described in this paper, an incremental approach that identifies subsystems of the complete system is an interesting avenue of investigation \cite{srinivasanKing08}.

\end{itemize}

\section{Summary and Conclusions}

In this paper we have presented a novel system, named ILP-QSI, which learns qualitative models of dynamic processes. This  system stands squarely in a strand of research that integrates Machine Learning with Qualitative Reasoning and extends the work in that area in the following ways:

The ILP-QSI algorithm itself extends the work; it is a branch and bound algorithm that makes use of background knowledge of (at least) three kinds in order to focus and guide the search for well posed models of dynamic processes.

\begin{description}

\item[Syntactic Constraints:] The model size is prespecified; models must be complete and determinate; and must not proliferate instances of qualitative relations.

\item[Semantic Constraints:] The model must adequately explain the data; it must not contain relations that are redundant or contradictory; and the relations in the model must respect dimensional constraints.

\item[System Theoretic Constraints:] The model should be singular and not disjoint; all endogenous variables must appear in at least two relations; and the model should be causally ordered.

\end{description}

We have thoroughly tested the system on a number of well known dynamic processes. This has enabled us to ascertain that ILP-QSI is capable of learning under a variety of conditions of noisy and noise free data. This testing has also allowed us to identify some conditions under which it is possible to learn an appropriate model of a dynamic system. The conclusions from this aspect of the work are:

\begin{itemize}

\item Learning precision is related to the richness (or sparcity) and noisiness of the data from which the learning is performed.

\item The target model is precisely learnt if the data used is a kernel subset.

\item These kernels are made up of states from different branches in the envisionment graph.

\item The system critical points play an important role in identifying the model structure.

\item There is a spectrum of possibilities with regard to the setting up of suitable experiments to garner data from which to learn models of the physical or biological systems of interest.

\item Cascaded parts of systems help to identify suitable points of decomposition for model learning.

\end{itemize}

While ILP-QSI is designed to learn a qualitative structural model from qualitative data, it is sometimes the case that the original measurements are quantitative (albeit sparse and possibly noisy). In order to ascertain how ILP-QSI would cope with qualitative data generated from quantitative measurements we carried out a ``proof-of-concept" set of experiments from each of the physical process models previously utilised. The results from these were in keeping with the results obtained from the qualitative experiments. This adds weight to the conclusions regarding the viability of our approach to learning structural models of dynamic systems under adverse conditions.

Finally, in order to test the scalability of the method, we applied ILP-QSI to a large scale metabolic pathway: glycolysis. In this case the search space was deemed too large to attempt learning the QSIM primitives alone. However, knowledge of the domain enabled us to group these primitives into a set of \textit{Metabolic Components} from which models of metabolic pathways can more easily be constructed. Also, for this part of the research Logical graph based models were used to represent background domain knowledge. Utilising these, we were able to identify 35 possible structures for the glycolysis pathway (out of a possible 27,254); of these the target model had the fewest cycles (though we do not know if this is a general phenomenon) and minimally covered the data generated from the reverse pathway of glucogenesis. 

The overall conclusions of the this work are that qualitative reasoning methods combined with machine learning (specifically ILP) can successfully learn qualitative structural models of systems of high complexity under a number of adverse circumstances. However, the work reported herein constitutes a step in a line of research that has only recently begun; and, as with all interesting lines of research, it raises in its turn interesting questions that need to be addressed.

\subsection*{Acknowledgments}

This work was supported in part by BBSRC/EPSRC grant BIO10479.  The authors would like to thank Stephen Oliver and Douglas Kell for their helpful discussions on the biological aspects of this paper. We would also like to thank Simon Garrett for many interesting and fruitful interactions. 

%\bibliographystyle{plain}
%\bibliography{alphaBib,biblio,mypubs,lrefs}

\appendix

\section{\label{solutionSpace} The Derivation of the Solution Space for the Tanks Systems}

In this appendix we provide a summary of whence the solution spaces for the tanks systems utilised in this project are constructed. Further details regarding envisionments and their associated solution spaces may be found in the work of \citeA{coghill92} and \citeA{coghill03}.

In order to facilitate this analysis we will need to make use of a quantitative version of the system models. For ease of exposition we will make the additional assumption that the systems are linear.\footnote{In fact for the types of non-linearity normally associated with systems of this kind the solution spaces are qualitatively identical to those described here, although the analysis required to construct them is slightly more complicated.}

\subsection{The U-tube}

A quantitative model of the u-tube system is

$$\frac{dh_1}{dt} =  k(h_1 - h_2)$$
$$\frac{dh_2}{dt} =  k(h_2 - h_1)$$

By inspection of these two equations it is easy to see that  (ignoring the trivial case where $k = 0$) the derivatives in these two equations are both zero when:

\begin{equation}h_1 = h_2\label{ute}\end{equation}

That is:

$$h_1 = h_2 \Rightarrow \frac{dh_1}{dt} = \frac{dh_2}{dt} = 0$$

This accounts for the relationship, depicted in Fig. \ref{tss}, between $h_1$ and $h_2$ when the derivatives are zero. From the envisionment table for the u-tube (Table \ref{utubenv} in Section \ref{qss}) we see that the only state with zero derivatives is state 5; hence it is represented by this line.

\subsection{The Coupled Tanks}

A quantitative model of the coupled tanks system is

\begin{equation} \frac{dh_1}{dt} = q_i - k_1(h_2 - h_1) \label{ce1}\end{equation}
\begin{equation}\frac{dh_2}{dt} = k_1(h_2 - h_1) - k_2 \cdot h_2 \label{ce2} \end{equation}

When $\frac{dh_1}{dt} =0$  Equation \ref{ce1} can be rewritten as:

$$0 = q_i - k_1(h_2 - h_1)$$
$$= q_i - k_1h_2 - k_1h_1$$

which can be re-arranged to give

$$h_2 = \frac{q_i}{k_1} - h_1$$

When  $q_i$ is zero this reduces to

\begin{equation}h_2 = h_1\label{ce3}\end{equation}

When $\frac{dh_2}{dt} =0$  Equation \ref{ce2} can be rewritten as:

$$0 = k_1(h_2 - h_1) - k_2h_2$$
$$= k_1h_1 - k_1h_1 - k_2h_2$$
$$= (k_1 - k_2)h_2 - k_1h_1$$

so

$$(k_1 - k_2)h_2 = k_1h_1$$

Re-arranging gives

\begin{equation}h_2 = \frac{k_1}{(k_1 - k_2)}h_1\label{ce4}\end{equation}

This accounts for the relations between $h_1$ and $h_2$ depicted in the solution space of Fig. \ref{tss}.

\subsection{The Cascaded Tanks}

A quantitative model of the cascaded tanks system is:

\begin{equation} \frac{dh_1}{dt} = q_i - k_1h_1 \label{ca1}\end{equation}
\begin{equation}\frac{dh_2}{dt} = k_1h_1 - k_2h_2 \label{ca2} \end{equation}

When $\frac{dh_1}{dt} =0$  Equation \ref{ca1} can be re-arranged as:

$$q_i = k_1h_1$$

or

$$h_1 =  \frac{q_i}{k_1}$$

When $\frac{dh_2}{dt} =0$  Equation \ref{ca2} can be rewritten as:

$$k_2h_2 =  k_1h_1$$

or 

$$h_2 =  \frac{k_1}{k_2}h_1$$

\begin{figure}[hbtp]
\begin{center}
\includegraphics[scale=0.65]{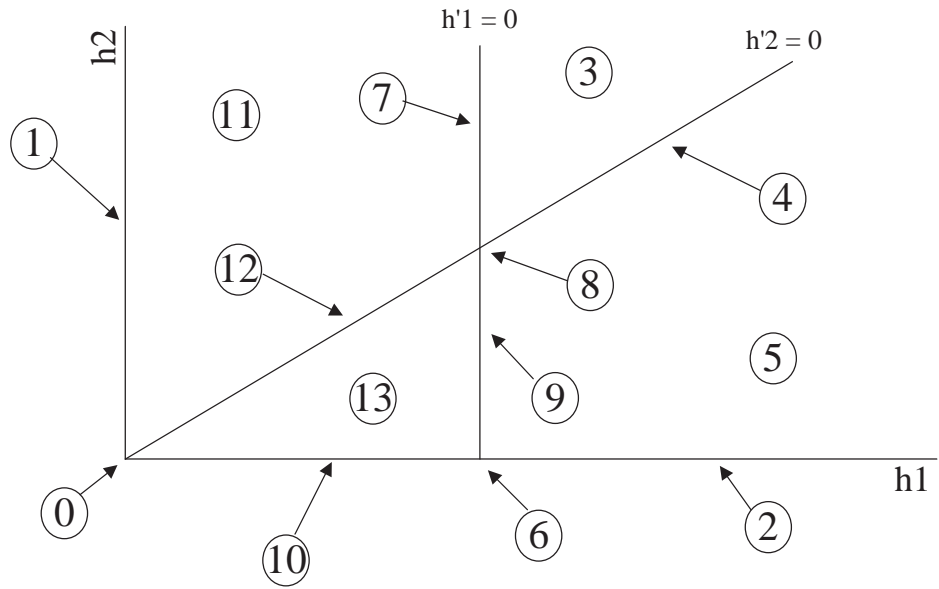}
\end{center}
\caption{\label{cassol} The solution space for the cascaded tanks system.}
\end{figure}

This accounts for the relations between $h_1$ and $h_2$ depicted in the solution space of Fig. \ref{cassol}.

%\newpage
%\bibliographystyle{plain}
\bibliography{jairBib}

\begin{thebibliography}{}

\bibitem[\protect\BCAY{Arkin, Shen,\ \BBA\ Ross}{Arkin
  et~al.}{1997}]{arkin:science}
Arkin, A., Shen, P., \BBA\ Ross, J. \BBOP1997\BBCP.
\newblock \BBOQ A test case of correlation metric construction of a reaction
  pathway from measurements\BBCQ\
\newblock {\Bem Science}, {\Bem 277}, 1275--1279.

\bibitem[\protect\BCAY{Asgharbeygi, Bay, Langley,\ \BBA\ Arrigo}{Asgharbeygi
  et~al.}{2006}]{asgharbeygi05}
Asgharbeygi, N., Bay, S., Langley, P., \BBA\ Arrigo, K. \BBOP2006\BBCP.
\newblock \BBOQ Inductive revision of quantitative process models\BBCQ\
\newblock {\Bem Ecological modelling}, {\Bem 194}, 70--79.

\bibitem[\protect\BCAY{Bergadano\ \BBA\ Gunetti}{Bergadano\ \BBA\
  Gunetti}{1996}]{BergadanoF96}
Bergadano, F.\BBACOMMA\  \BBA\ Gunetti, D. \BBOP1996\BBCP.
\newblock {\Bem Inductive Logic Programming: From Machine Learning to Software
  Engineering}.
\newblock MIT Press.

\bibitem[\protect\BCAY{Bhaskhar\ \BBA\ Nigam}{Bhaskhar\ \BBA\
  Nigam}{1990}]{Bhaskar90}
Bhaskhar, R.\BBACOMMA\  \BBA\ Nigam, A. \BBOP1990\BBCP.
\newblock \BBOQ Qualitative physics using dimensional analysis\BBCQ\
\newblock {\Bem Artificial Intelligence}, {\Bem 45}, 73--111.

\bibitem[\protect\BCAY{Blackman\ \BBA\ Tukey}{Blackman\ \BBA\
  Tukey}{1958}]{Blackman58}
Blackman, R.~B.\BBACOMMA\  \BBA\ Tukey, J.~W. \BBOP1958\BBCP.
\newblock {\Bem The Measurement of Power Spectra}.
\newblock John Wiley and Sons, New York.

\bibitem[\protect\BCAY{Bradley, Easley,\ \BBA\ Stolle}{Bradley
  et~al.}{2000}]{Bradley00}
Bradley, E., Easley, M., \BBA\ Stolle, R. \BBOP2000\BBCP.
\newblock \BBOQ Reasoning about nonlinear system identification\BBCQ\
\newblock \BTR\ CU-CS-894-99, University of Colorado.

\bibitem[\protect\BCAY{Bratko, Mozetic,\ \BBA\ Lavrac}{Bratko
  et~al.}{1989}]{kardio}
Bratko, I., Mozetic, I., \BBA\ Lavrac, N. \BBOP1989\BBCP.
\newblock {\Bem KARDIO: A Study in Deep and Qualitative Knowledge for Expert
  Systems}.
\newblock MIT Press, Cambridge, Massachusetts.

\bibitem[\protect\BCAY{Bratko, Muggleton,\ \BBA\ Varsek}{Bratko
  et~al.}{1992}]{Muggleton91}
Bratko, I., Muggleton, S., \BBA\ Varsek, A. \BBOP1992\BBCP.
\newblock \BBOQ Learning qualitative models of dynamic systems\BBCQ\
\newblock In Muggleton, S.\BED, {\Bem Inductive Logic Programming}, \BPGS\
  437--452. Academic Press.

\bibitem[\protect\BCAY{Bridewell, Sandy,\ \BBA\ Langley}{Bridewell
  et~al.}{2004}]{bridewell04}
Bridewell, W., Sandy, J., \BBA\ Langley, P. \BBOP2004\BBCP.
\newblock \BBOQ An interactive environment for the modeling and discovery of
  scientific knowledge.\BBCQ\
\newblock \BTR, Institute for the Study of Learning and Expertise, Palo Alto,
  CA.

\bibitem[\protect\BCAY{Camacho}{Camacho}{2000}]{rui:thesis}
Camacho, R. \BBOP2000\BBCP.
\newblock {\Bem Inducing Models of Human Control Skills using Machine Learning
  Algorithms}.
\newblock Ph.D.\ thesis, University of Porto.

\bibitem[\protect\BCAY{Coghill}{Coghill}{1996}]{Coghill96}
Coghill, G.~M. \BBOP1996\BBCP.
\newblock {\Bem Mycroft: A Framework for Constraint-Based Fuzzy Qualitative
  Reasoning}.
\newblock Ph.D.\ thesis, Heriot-Watt University.

\bibitem[\protect\BCAY{Coghill}{Coghill}{2003}]{coghill03}
Coghill, G.~M. \BBOP2003\BBCP.
\newblock \BBOQ Fuzzy envisionment\BBCQ\
\newblock In {\Bem Proc. of the Third International Workshop on Hybrid Methods
  for Adaptive Systems}, Oulu, Finland.

\bibitem[\protect\BCAY{Coghill, Asbury, van Rijsbergen,\ \BBA\ Gray}{Coghill
  et~al.}{1992}]{coghill92}
Coghill, G.~M., Asbury, A.~J., van Rijsbergen, C.~J., \BBA\ Gray, W.~M.
  \BBOP1992\BBCP.
\newblock \BBOQ The application of vector envisionment to compartmental
  systems.\BBCQ\
\newblock In {\Bem Proceedings of the first international conference on
  Intelligent Systems Engineering}, \BPGS\ 123--128, Edinburgh, Scotland.

\bibitem[\protect\BCAY{Coghill\ \BBA\ Chantler}{Coghill\ \BBA\
  Chantler}{1994}]{Coghill94}
Coghill, G.~M.\BBACOMMA\  \BBA\ Chantler, M.~J. \BBOP1994\BBCP.
\newblock \BBOQ Mycroft: a framework for qualitative reasoning\BBCQ\
\newblock In {\Bem Proceedings of the Second International Conference on
  Intelligent Systems Engineering}, \BPGS\ 49--55, Hamburg, Germany.

\bibitem[\protect\BCAY{Coghill, Garret,\ \BBA\ King}{Coghill
  et~al.}{2004}]{coghill04}
Coghill, G.~M., Garret, S.~M., \BBA\ King, R.~D. \BBOP2004\BBCP.
\newblock \BBOQ Learning qualitative models of metabolic systems\BBCQ\
\newblock In {\Bem Proceedings of the European Conference on Artificial
  Intelligence ECAI-04}, Valencia, Spain.

\bibitem[\protect\BCAY{Coghill\ \BBA\ Shen}{Coghill\ \BBA\
  Shen}{2001}]{Coghill01}
Coghill, G.~M.\BBACOMMA\  \BBA\ Shen, Q. \BBOP2001\BBCP.
\newblock \BBOQ On the specification of multiple models for diagnosis of
  dynamic systems\BBCQ\
\newblock {\Bem AI Communications}, {\Bem 14\/}(2), 93--104.

\bibitem[\protect\BCAY{Coiera}{Coiera}{1989a}]{Coiera89}
Coiera, E.~W. \BBOP1989a\BBCP.
\newblock \BBOQ Generating qualitative models from example behaviours\BBCQ\
\newblock \BTR\ 8901, University of New South Wales, Deptartment of Computer
  Science.

\bibitem[\protect\BCAY{Coiera}{Coiera}{1989b}]{Coiera89a}
Coiera, E.~W. \BBOP1989b\BBCP.
\newblock \BBOQ Learning qualitative models from example behaviours\BBCQ\
\newblock In {\Bem Proc. Third Workshop on Qualitative Physics}, Stanford.

\bibitem[\protect\BCAY{Collins}{Collins}{1968}]{coll:mi2}
Collins, J. \BBOP1968\BBCP.
\newblock \BBOQ A regression analysis program incorporating heuristic term
  selection\BBCQ\
\newblock In Dale, E.\BBACOMMA\  \BBA\ Michie, D.\BEDS, {\Bem Machine
  Intelligence 2}. Oliver and Boyd.

\bibitem[\protect\BCAY{D{\v z}eroski}{D{\v z}eroski}{1992}]{Dzeroski92}
D{\v z}eroski, S. \BBOP1992\BBCP.
\newblock \BBOQ Learning qualitative models with inductive logic
  programming\BBCQ\
\newblock {\Bem Informatica}, {\Bem 16\/}(4), 30--41.

\bibitem[\protect\BCAY{D{\v z}eroski\ \BBA\ Todorovski}{D{\v z}eroski\ \BBA\
  Todorovski}{1993}]{Dzeroski93}
D{\v z}eroski, S.\BBACOMMA\  \BBA\ Todorovski, L. \BBOP1993\BBCP.
\newblock \BBOQ Discovering dynamics\BBCQ\
\newblock In {\Bem International Conference on Machine Learning}, \BPGS\
  97--103.

\bibitem[\protect\BCAY{D{\v z}eroski\ \BBA\ Todorovski}{D{\v z}eroski\ \BBA\
  Todorovski}{1995}]{Dzeroski95}
D{\v z}eroski, S.\BBACOMMA\  \BBA\ Todorovski, L. \BBOP1995\BBCP.
\newblock \BBOQ Discovering dynamics: from inductive logic programming to
  machine discovery\BBCQ\
\newblock {\Bem J. Intell. Information Syst.}, {\Bem 4}, 89--108.

\bibitem[\protect\BCAY{Falkenhainer\ \BBA\ Michalski}{Falkenhainer\ \BBA\
  Michalski}{1986}]{falkenhainer:abacus}
Falkenhainer, B.\BBACOMMA\  \BBA\ Michalski, R.~S. \BBOP1986\BBCP.
\newblock \BBOQ Integrating quantitative and qualitative discovery: The abacus
  system\BBCQ\
\newblock {\Bem Machine Learning}, {\Bem 1\/}(4), 367--401.

\bibitem[\protect\BCAY{Forbus}{Forbus}{1984}]{Forbus84}
Forbus, K.~D. \BBOP1984\BBCP.
\newblock \BBOQ Qualitative process theory\BBCQ\
\newblock {\Bem Artificial Intelligence}, {\Bem 24}, 169--204.

\bibitem[\protect\BCAY{Garrett, Coghill, Srinivasan,\ \BBA\ King}{Garrett
  et~al.}{2007}]{garrett07}
Garrett, S.~M., Coghill, G.~M., Srinivasan, A., \BBA\ King, R.~D.
  \BBOP2007\BBCP.
\newblock \BBOQ Learning qualitative models of physical and biological
  systems\BBCQ\
\newblock In D{\v z}eroski, S.\BBACOMMA\  \BBA\ Todorovski, L.\BEDS, {\Bem
  Computational discovery of communicable knowledge}, \BPGS\ 248--272.
  Springer.

\bibitem[\protect\BCAY{Gawthrop\ \BBA\ Smith}{Gawthrop\ \BBA\
  Smith}{1996}]{Gawthrop96}
Gawthrop, P.~J.\BBACOMMA\  \BBA\ Smith, L. P.~S. \BBOP1996\BBCP.
\newblock {\Bem Metamodelling: Bond Graphs and Dynamic Systems}.
\newblock Prentice Hall, Hemel Hempstead, Herts, England.

\bibitem[\protect\BCAY{Hau\ \BBA\ Coiera}{Hau\ \BBA\ Coiera}{1993}]{Hau93}
Hau, D.~T.\BBACOMMA\  \BBA\ Coiera, E.~W. \BBOP1993\BBCP.
\newblock \BBOQ Learning qualitative models of dynamic systems\BBCQ\
\newblock {\Bem Machine Learning}, {\Bem 26}, 177--211.

\bibitem[\protect\BCAY{Healey}{Healey}{1975}]{Healey75}
Healey, M. \BBOP1975\BBCP.
\newblock {\Bem Principles of Automatic Control}.
\newblock Hodder and Stoughton.

\bibitem[\protect\BCAY{Iwasaki\ \BBA\ Simon}{Iwasaki\ \BBA\
  Simon}{1986}]{Iwasaki86}
Iwasaki, Y.\BBACOMMA\  \BBA\ Simon, H.~A. \BBOP1986\BBCP.
\newblock \BBOQ Causality in device behavior\BBCQ\
\newblock {\Bem Artificial Intelligence}, {\Bem 29}, 3--32.
\newblock See also {De Kleer} and Brown's rebuttal and Iwasaki and Simon's
  reply to their rebuttal in the same volume of this journal.

\bibitem[\protect\BCAY{King, Garrett,\ \BBA\ Coghill}{King
  et~al.}{2005}]{king05}
King, R.~D., Garrett, S.~M., \BBA\ Coghill, G.~M. \BBOP2005\BBCP.
\newblock \BBOQ On the use of qualitative reasoning to simulate and identify
  metabolic pathways.\BBCQ\
\newblock {\Bem Bioinformatics}, {\Bem 21\/}(9), 2017 -- 2026.

\bibitem[\protect\BCAY{Kokar}{Kokar}{1985}]{kokar:coper}
Kokar, M.~M. \BBOP1985\BBCP.
\newblock \BBOQ Coper: A methodology for learning invariant functional
  descriptions\BBCQ\
\newblock In Mitchell, T., Carbonell, J., \BBA\ Michalski, R.\BEDS, {\Bem
  Machine Learning: A Guide to Current Research}, \BPGS\ 151--154. Kluwer
  Academic Press.

\bibitem[\protect\BCAY{Koza, Mydlowec, Lanza, Yu,\ \BBA\ Keane}{Koza
  et~al.}{2000}]{kosa:pathway}
Koza, J.~R., Mydlowec, W., Lanza, G., Yu, J., \BBA\ Keane, M.~A.
  \BBOP2000\BBCP.
\newblock \BBOQ Reverse engineering and automatic synthesis of metabolic
  pathways from observed data using genetic programming.\BBCQ\
\newblock \BTR\ SMI-2000-0851, Stanford University.

\bibitem[\protect\BCAY{Kraan, Richards,\ \BBA\ Kuipers}{Kraan
  et~al.}{1991}]{Kraan91}
Kraan, I.~C., Richards, B.~L., \BBA\ Kuipers, B.~J. \BBOP1991\BBCP.
\newblock \BBOQ Automatic abduction of qualitative models\BBCQ\
\newblock In {\Bem Proceedings of Qualitative Reasoning 1991 (QR'91)}.

\bibitem[\protect\BCAY{Kuipers}{Kuipers}{1994}]{Kuipers2}
Kuipers, B. \BBOP1994\BBCP.
\newblock {\Bem Qualitative Reasoning}.
\newblock MIT Press.

\bibitem[\protect\BCAY{Langley}{Langley}{1981}]{LangleyP81}
Langley, P. \BBOP1981\BBCP.
\newblock \BBOQ Data-driven discovery of physical laws\BBCQ\
\newblock {\Bem Cognitive Science}, {\Bem 5}, 31--54.

\bibitem[\protect\BCAY{Langley, George, Bay,\ \BBA\ Saito}{Langley
  et~al.}{2003}]{langley03}
Langley, P., George, D., Bay, S., \BBA\ Saito, K. \BBOP2003\BBCP.
\newblock \BBOQ Robust induction of process models from time series data.\BBCQ\
\newblock In {\Bem Proc. twentieth International Conference on Machine
  Learning}, \BPGS\ 432--439, Washington, DC. AAAI Press.

\bibitem[\protect\BCAY{Langley\ \BBA\ Zytkow}{Langley\ \BBA\
  Zytkow}{1989}]{langley:fahrenheit}
Langley, P.\BBACOMMA\  \BBA\ Zytkow, J. \BBOP1989\BBCP.
\newblock \BBOQ Data-driven approaches to empirical discovery\BBCQ\
\newblock {\Bem Artificial Intelligence}, {\Bem 40}, 283--312.

\bibitem[\protect\BCAY{McCreath}{McCreath}{1999}]{mccreath:thesis}
McCreath, E. \BBOP1999\BBCP.
\newblock {\Bem Induction in first order logic from noisy training examples and
  fixed example set sizes}.
\newblock Ph.D.\ thesis, University of New South Wales.

\bibitem[\protect\BCAY{Mitchell, Keller,\ \BBA\ Kedar-Cabelli}{Mitchell
  et~al.}{1986}]{MitchellT86}
Mitchell, T.~M., Keller, R.~M., \BBA\ Kedar-Cabelli, S. \BBOP1986\BBCP.
\newblock \BBOQ Explanation-based generalization: A unifying view\BBCQ\
\newblock {\Bem Machine Learning}, {\Bem 1}, 47--80.

\bibitem[\protect\BCAY{Mozetic}{Mozetic}{1987}]{Mozetic87}
Mozetic, I. \BBOP1987\BBCP.
\newblock \BBOQ Learning of qualitative models\BBCQ\
\newblock In Bratko, I.\BBACOMMA\  \BBA\ Lavrac, N.\BEDS, {\Bem Progress in
  Machine Learning: Proceedings of EWSL '87: 2nd European Working Session on
  Learning}, \BPGS\ 201--217. Sigma Press.

\bibitem[\protect\BCAY{Muggleton}{Muggleton}{1995}]{mugg:progol}
Muggleton, S. \BBOP1995\BBCP.
\newblock \BBOQ {I}nverse {E}ntailment and {P}rogol\BBCQ\
\newblock {\Bem New Gen. Comput.}, {\Bem 13}, 245--286.

\bibitem[\protect\BCAY{Muggleton}{Muggleton}{1996}]{mugg:poslearn}
Muggleton, S. \BBOP1996\BBCP.
\newblock \BBOQ Learning from positive data\BBCQ\
\newblock {\Bem Lecture Notes in AI}, {\Bem 1314}, 358--376.

\bibitem[\protect\BCAY{Muggleton\ \BBA\ Feng}{Muggleton\ \BBA\
  Feng}{1990}]{Muggleton90}
Muggleton, S.\BBACOMMA\  \BBA\ Feng, C. \BBOP1990\BBCP.
\newblock \BBOQ Efficient induction of logic programs\BBCQ\
\newblock In {\Bem Proc. of the First Conf. on Algorithmic Learning Theory}.
  OHMSHA, Tokyo.

\bibitem[\protect\BCAY{Muggleton\ \BBA\ Raedt}{Muggleton\ \BBA\
  Raedt}{1994}]{mugg:der}
Muggleton, S.\BBACOMMA\  \BBA\ Raedt, L.~D. \BBOP1994\BBCP.
\newblock \BBOQ Inductive logic programming: Theory and methods\BBCQ\
\newblock {\Bem Journal of Logic Programming}, {\Bem 19,20}, 629--679.

\bibitem[\protect\BCAY{Narasimhan, Mosterman,\ \BBA\ Biswas}{Narasimhan
  et~al.}{1998}]{biswas98}
Narasimhan, S., Mosterman, P., \BBA\ Biswas, G. \BBOP1998\BBCP.
\newblock \BBOQ A systematic analysis of measurement selection algorithms for
  fault isolation in dynamic systems\BBCQ\
\newblock In {\Bem Proc. Ninth Intl. Workshop on Principles of Diagnosis
  (DX-98)}, \BPGS\ 94--101, Cape Cod, MA.

\bibitem[\protect\BCAY{Nordhausen\ \BBA\ Langley}{Nordhausen\ \BBA\
  Langley}{1993}]{Nordhausen93}
Nordhausen, B.\BBACOMMA\  \BBA\ Langley, P. \BBOP1993\BBCP.
\newblock \BBOQ An integrated framework for empirical discovery\BBCQ\
\newblock {\Bem Machine Learning}, {\Bem 12}, 17--47.

\bibitem[\protect\BCAY{Papadimitriou\ \BBA\ Steiglitz}{Papadimitriou\ \BBA\
  Steiglitz}{1982}]{pap:optim}
Papadimitriou, C.\BBACOMMA\  \BBA\ Steiglitz, K. \BBOP1982\BBCP.
\newblock {\Bem Combinatorial {O}ptimisation}.
\newblock Prentice-Hall, Edgewood-Cliffs, NJ.

\bibitem[\protect\BCAY{Plotkin}{Plotkin}{1971}]{plotkin:thesis}
Plotkin, G. \BBOP1971\BBCP.
\newblock {\Bem Automatic Methods of Inductive Inference}.
\newblock Ph.D.\ thesis, Edinburgh University.

\bibitem[\protect\BCAY{Reiser, King, Kell, Muggleton, Bryant,\ \BBA\
  Oliver}{Reiser et~al.}{2001}]{reiser:yeast}
Reiser, P., King, R., Kell, D., Muggleton, S., Bryant, C., \BBA\ Oliver, S.
  \BBOP2001\BBCP.
\newblock \BBOQ Developing a logical model of yeast metabolism\BBCQ\
\newblock {\Bem Electronic Transactions on Artificial Intelligence}, {\Bem 5},
  233--244.

\bibitem[\protect\BCAY{Richards, Kraan,\ \BBA\ Kuipers}{Richards
  et~al.}{1992}]{richards:qm}
Richards, B.~L., Kraan, I., \BBA\ Kuipers, B.~J. \BBOP1992\BBCP.
\newblock \BBOQ Automatic abduction of qualitative models\BBCQ\
\newblock In {\Bem Proc. of the Tenth National Conference on Artificial
  Intelligence (AAAI-92)}, \BPGS\ 723--728. MIT Press.

\bibitem[\protect\BCAY{Richards\ \BBA\ Mooney}{Richards\ \BBA\
  Mooney}{1995}]{forte}
Richards, B.~L.\BBACOMMA\  \BBA\ Mooney, R.~J. \BBOP1995\BBCP.
\newblock \BBOQ Automated refinement of first-order horn-clause domain
  theories\BBCQ\
\newblock {\Bem Machine Learning}, {\Bem 19\/}(2), 95--131.

\bibitem[\protect\BCAY{Riguzzi}{Riguzzi}{2005}]{riguzzi:ilp2005}
Riguzzi, F. \BBOP2005\BBCP.
\newblock \BBOQ Two results regarding refinement operators\BBCQ\
\newblock In Kramer, S.\BBACOMMA\  \BBA\ Pfahringer, B.\BEDS, {\Bem Late
  Breaking Papers, 15th International Workshop on Inductive Logic Programming
  ({ILP}05), August 10--13, 2005}, \BPGS\ 53--58, M\"{u}nich, Germany.

\bibitem[\protect\BCAY{Sadot, Fisher, Barak, Admanit, Stern, Hubbard,\ \BBA\
  Harel}{Sadot et~al.}{2008}]{sadot.et.al.08}
Sadot, A., Fisher, J., Barak, D., Admanit, Y., Stern, M.~J., Hubbard, E. J.~A.,
  \BBA\ Harel, D. \BBOP2008\BBCP.
\newblock \BBOQ Towards verified biological models.\BBCQ\
\newblock {\Bem IEEE/ACM Trans. Comput. Biology and Bioinformatics.}, {\Bem
  5(2)}, 1--12.

\bibitem[\protect\BCAY{Say\ \BBA\ Kuru}{Say\ \BBA\ Kuru}{1996}]{SayKuru96}
Say, A. C.~C.\BBACOMMA\  \BBA\ Kuru, S. \BBOP1996\BBCP.
\newblock \BBOQ Qualitative system identification: deriving structure from
  behavior\BBCQ\
\newblock {\Bem Artificial Intelligence}, {\Bem 83}, 75--141.

\bibitem[\protect\BCAY{Shoup}{Shoup}{1979}]{Shoup79}
Shoup, T.~E. \BBOP1979\BBCP.
\newblock {\Bem A Practical Guide to Computer Methods for Engineers}.
\newblock Prentice-Hall Inc., Englewood Cliffs, N.\,J.\,07632.

\bibitem[\protect\BCAY{Soderstrom\ \BBA\ Stoica}{Soderstrom\ \BBA\
  Stoica}{1989}]{soderstrom:sysid}
Soderstrom, T.\BBACOMMA\  \BBA\ Stoica, P. \BBOP1989\BBCP.
\newblock {\Bem System Identification}.
\newblock Prentice Hall.

\bibitem[\protect\BCAY{Srinivasan}{Srinivasan}{1999}]{aleph}
Srinivasan, A. \BBOP1999\BBCP.
\newblock \BBOQ {The Aleph Manual}\BBCQ\
\newblock Available at http://www.comlab.ox.ac.uk/oucl/
  research/areas/machlearn/Aleph/.

\bibitem[\protect\BCAY{Srinivasan\ \BBA\ King}{Srinivasan\ \BBA\
  King}{2008}]{srinivasanKing08}
Srinivasan, A.\BBACOMMA\  \BBA\ King, R.~D. \BBOP2008\BBCP.
\newblock \BBOQ Incremental identification of qualitative models of biological
  systems using inductive logic programming\BBCQ\
\newblock {\Bem J. Machine Learning Research {\em to appear}}.

\bibitem[\protect\BCAY{Todorovski}{Todorovski}{2003}]{Todorovski03}
Todorovski, L. \BBOP2003\BBCP.
\newblock {\Bem Using domain knowledge for automated modeling of dynamic
  systems with equation discovery}.
\newblock Ph.D.\ thesis, Faculty of Electrical Engineering and Computer
  Science, University of Ljubljana, Slovenia.

\bibitem[\protect\BCAY{Todorovski, Srinivasan, Whiteley,\ \BBA\
  Gavaghan}{Todorovski et~al.}{2000}]{Todorovski00}
Todorovski, L., Srinivasan, A., Whiteley, J., \BBA\ Gavaghan, D.
  \BBOP2000\BBCP.
\newblock \BBOQ Discovering the structure of partial differential equations
  from example behavior\BBCQ\
\newblock In {\Bem Proceedings of the Seventeenth International Conference on
  Machine Learning}, \BPGS\ 991--998, San Francisco.

\bibitem[\protect\BCAY{Todorovski\ \BBA\ D{\v z}eroski}{Todorovski\ \BBA\ D{\v
  z}eroski}{1997}]{Todorovski97}
Todorovski, L.\BBACOMMA\  \BBA\ D{\v z}eroski, S. \BBOP1997\BBCP.
\newblock \BBOQ Declarative bias in equation discovery\BBCQ\
\newblock In {\Bem Proc. 14th International Conference on Machine Learning},
  \BPGS\ 376--384. Morgan Kaufmann.

\bibitem[\protect\BCAY{Valdes-Perez}{Valdes-Perez}{1994}]{valdesPerez94}
Valdes-Perez, R.~E. \BBOP1994\BBCP.
\newblock \BBOQ Heuristics for systematic elucidation of reaction
  pathways.\BBCQ\
\newblock {\Bem J. Chem. Informat. Comput. Sci.}, {\Bem 34}, 976--983.

\bibitem[\protect\BCAY{Voit\ \BBA\ Radivoyevitch}{Voit\ \BBA\
  Radivoyevitch}{2000}]{voit00}
Voit, E.~O.\BBACOMMA\  \BBA\ Radivoyevitch, T. \BBOP2000\BBCP.
\newblock \BBOQ Biochemical systems analysis of genome-wide expression
  data\BBCQ\
\newblock {\Bem Bioinformatics}, {\Bem 16\/}(11), 1023--1037.

\bibitem[\protect\BCAY{Warren, Coghill,\ \BBA\ Johnstone}{Warren
  et~al.}{2004}]{warren04}
Warren, P., Coghill, G.~M., \BBA\ Johnstone, A. \BBOP2004\BBCP.
\newblock \BBOQ Top down and botton up development of a fuzzy rule-based
  diagnostic system\BBCQ\
\newblock In {\Bem Proc. of the Fourth International Workshop on Hybrid Methods
  for Adaptive Systems}, Aachen, Germany.

\end{thebibliography}
\bibliographystyle{theapa}
%\bibliography{alphaBib,biblio,mypubs,lrefs}

\end{document}